%% file: main_arxiv.tex
\documentclass[10pt,twocolumn,letterpaper]{article}

\usepackage[pagenumbers]{cvpr}
\usepackage[accsupp]{axessibility}  %
\input{preamble}

\usepackage[pagebackref,breaklinks,colorlinks]{hyperref}

\def\paperID{*****} %
\def\confName{CVPR}
\def\confYear{2025}

\title{Tiled Diffusion}

\author{Or Madar\\
Reichman University\\
{\tt\small or.madar@post.runi.ac.il}
\and
Ohad Fried\\
Reichman University\\
{\tt\small ofried@runi.ac.il}
}

\input{macros}

\begin{document}

\twocolumn[{%
\renewcommand\twocolumn[1][]{#1}%
\maketitle

\input{figures/teaser/teaser}
}]
\input{sec/0_abstract}

\textcolor{blue}{For the supplemental material and code we refer the readers to our paper  \href{https://madaror.github.io/tiled-diffusion.github.io/}{webpage}}
\input{sec/1_intro}

\input{sec/2_related}

\input{sec/3_method}

\input{sec/4_evaluation}

\input{sec/5_applications}
\input{sec/6_limitations}
\input{sec/7_conclusion}
\input{sec/8_ack}

{
    \small
    \bibliographystyle{ieeenat_fullname}
    \bibliography{main}
}

\end{document}

%% file: preamble.tex
\usepackage[export]{adjustbox}
\usepackage{xcolor,colortbl}
\usepackage{multirow}
\usepackage{pifont}
\usepackage{float}
\usepackage{svg}
\usepackage{caption}
\usepackage{multicol}
\usepackage{tikz}
\usepackage{relsize}
\usepackage[outline]{contour}
\contourlength{1.2pt}
\usepackage{amssymb}    %
\usepackage{graphicx}
\usepackage{amsmath}
\usepackage{booktabs}

%% file: macros.tex
\newcommand{\ignorethis}[1]{}

\newcommand{\Reals      }     {{\textrm{I\kern-0.18em R}}}

\newcommand{\change     } [1] {\mbox{{\footnotesize $\Delta$} \kern-3pt}#1}

\definecolor{darkred}{rgb}{0.7,0.1,0.1}
\definecolor{darkgreen}{rgb}{0.1,0.6,0.1}
\definecolor{cyan}{rgb}{0.7,0.0,0.7}
\definecolor{otherblue}{rgb}{0.1,0.4,0.8}
\definecolor{maroon}{rgb}{0.76,.13,.28}
\definecolor{burntorange}{rgb}{0.81,.33,0}

\newif\ifdraft
\drafttrue

\ifdraft
  \newcommand{\orm}[1]{{\color{burntorange}[\textbf{Or:} #1]}}
  \newcommand{\ohad}[1]{{\color{magenta}[\textbf{Ohad:} #1]}}

\else
  \newcommand{\orm}[1]{}
  \newcommand{\ohad}[1]{}

\fi

\newcommand\todosilent[1]{}

\definecolor{colora}{RGB}{242,78,112}

\definecolor{colorb}{RGB}{255,176,1}

\definecolor{colorc}{RGB}{41,160,177}

\definecolor{colord}{RGB}{55, 146, 55}

\newcommand{\xmark}{\color{darkred}\ding{55}}%

%% file: figures/teaser/teaser.tex
\begin{center}
    \centering
    \captionsetup{type=figure}
    \includegraphics[width=\textwidth]{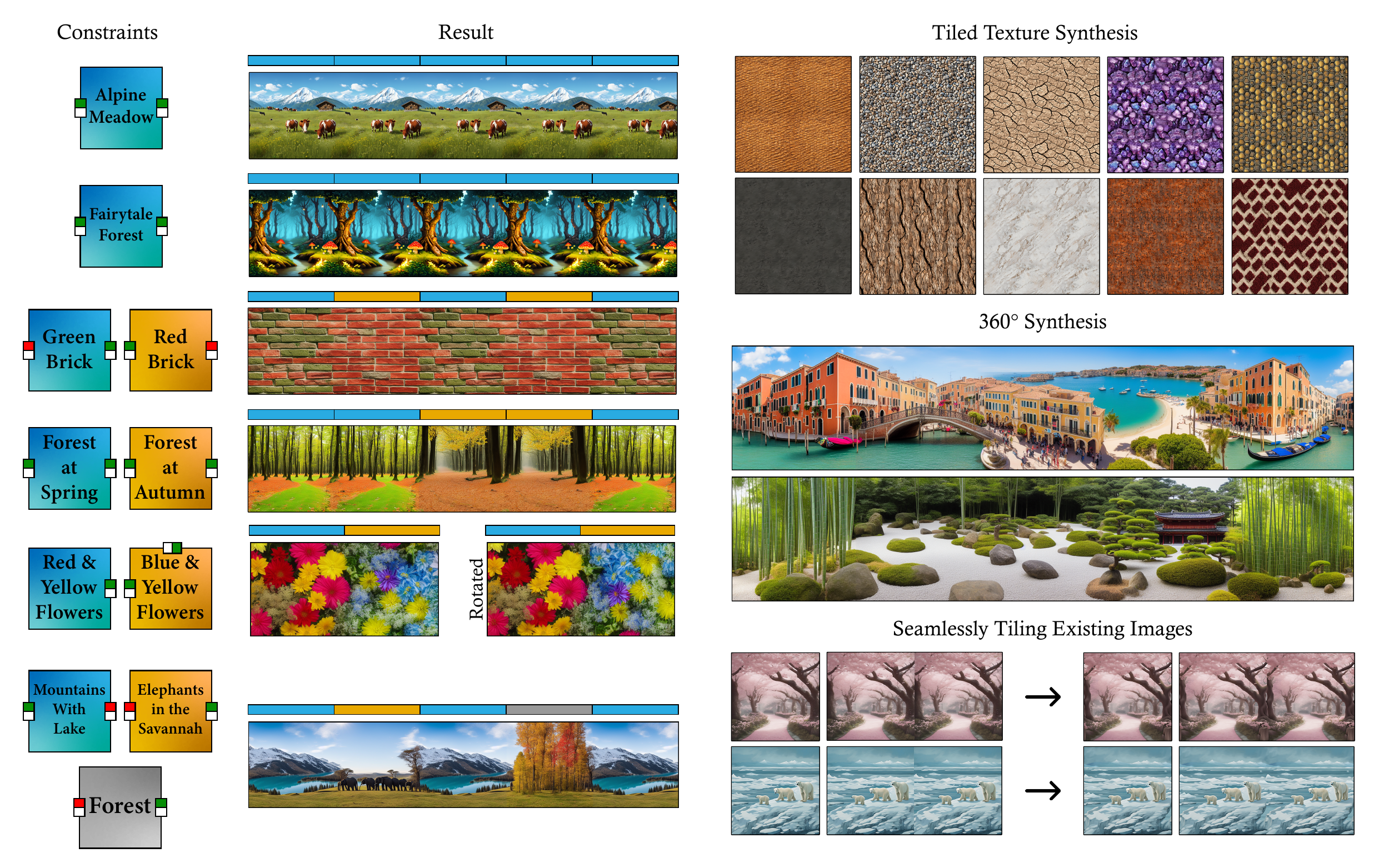}
\captionof{figure}{
\textbf{Tiled Diffusion.} Our method generates seamlessly tileable images using diffusion models.
Left: input constraints and their results. 
Matching color patterns on edges indicate they should tile seamlessly. (We use two colors to convey the direction.)
Color strip above the result shows which constraint was used for each area.
Right: various applications of our method: tiled texture synthesis, 360\textdegree{} synthesis, and seamlessly tiling existing images. 
Each texture includes 4 tiles (2x2); the 360\textdegree{} examples wrap on the horizontal axis; the seamlessly tiling example shows original images (left) and their tileable versions (right), both in 1x2 arrangements.
}
\label{teaser}
\end{center}

%% file: sec/0_abstract.tex
\begin{abstract}
Image tiling---the seamless connection of disparate images to create a coherent visual field---is crucial for applications such as texture creation, video game asset development, and digital art. Traditionally, tiles have been constructed manually, a method that poses significant limitations in scalability and flexibility. Recent research has attempted to automate this process using generative models. However, current approaches primarily focus on tiling textures and manipulating models for single-image generation, without inherently supporting the creation of multiple interconnected tiles across diverse domains.
This paper presents Tiled Diffusion, a novel approach that extends the capabilities of diffusion models to accommodate the generation of cohesive tiling patterns across various domains of image synthesis that require tiling. Our method supports a wide range of tiling scenarios, from self-tiling to complex many-to-many connections, enabling seamless integration of multiple images.
Tiled Diffusion automates the tiling process, eliminating the need for manual intervention and enhancing creative possibilities in various applications, such as seamlessly tiling of existing images, tiled texture creation, and 360° synthesis.
\end{abstract}

%% file: sec/1_intro.tex
\section{Introduction}
\label{sec:intro}
Tiling, the seamless connection of images to create coherent visual fields, is crucial in digital image processing and generation. It is essential for applications ranging from texture creation to video game asset development and digital art. Traditionally, manual tile creation has limited scalability and flexibility. Recent research \cite{controlmat,Rodriguez-Pardo_2024_CVPR} has attempted to automate this process using generative models, but current approaches primarily focus on simple tiling scenarios, mainly single image generation in the field of texture synthesis.
We define the tiling problem using a list of constraints. We consider each image $I_i$ as a rectangle with four sides (right ($I_{i|}$), left ($I_{|i}$), top ($I_{\overline{i}}$), and bottom ($I_{\underline{i}}$)) and each constraint $C_j$ is defined by two sets of sides. All the sides from one set can connect to all of the sides of the other set. These sets can include sides from any of the images being tiled. This flexible definition allows for various tiling scenarios. Examples of these scenarios are illustrated in \Cref{intro:example}:

\begin{itemize}
\item Self-tiling: For single images that repeat seamlessly. As shown in \Cref{intro:example}~(left), $I_1$ connects only to itself. Formally, we have two constraints:
    $C_1 = \{\{I_{1|}\}, \{I_{|1}\}\}$ and $C_2 = \{\{I_{\overline{1}}\}, \{I_{\underline{1}}\}\}$.
\item One-to-one tiling: Allows each side of an image to connect to a maximum of one side of another image. \Cref{intro:example}~(middle) demonstrates this for two images on the X-axis. Each constraint's sets contain exactly one element. %
\item Many-to-many tiling: Supports any number of connections per side, including complex cross-axis connections (X-axis to Y-axis). \Cref{intro:example}~(right) shows a many-to-many scenario where right sides of both images can connect to left sides of both images. There are no limitations on the number of sides in either set.
\end{itemize}

Our method, Tiled Diffusion, extends the capabilities of diffusion models to accommodate these diverse tiling scenarios. It addresses several challenges, including ensuring content coherence across diverse tile boundaries, maintaining stylistic consistency between connected images, and preserving both local details and global structure in complex arrangements.
We introduce two key innovations: a tiling constraint for global structure consistency and local detail matching, and a similarity constraint to eliminate artifacts in complex tiling scenarios.
We demonstrate our method's effectiveness in several applications which are seamlessly tiling existing images, tileable texture generation and 360° synthesis.

\input{figures/intro/intro}

%% file: figures/intro/intro.tex
\newlength{\intro}
\setlength{\intro}{1.4cm}

\begin{figure}[t]

    \centering
    \setlength{\tabcolsep}{3pt} %

    \begin{tabular}{ccccc}
        Self-tiling & \multicolumn{2}{c}{One-to-one} & \multicolumn{2}{c}{Many-to-many} \\
        \includegraphics[height=\intro]{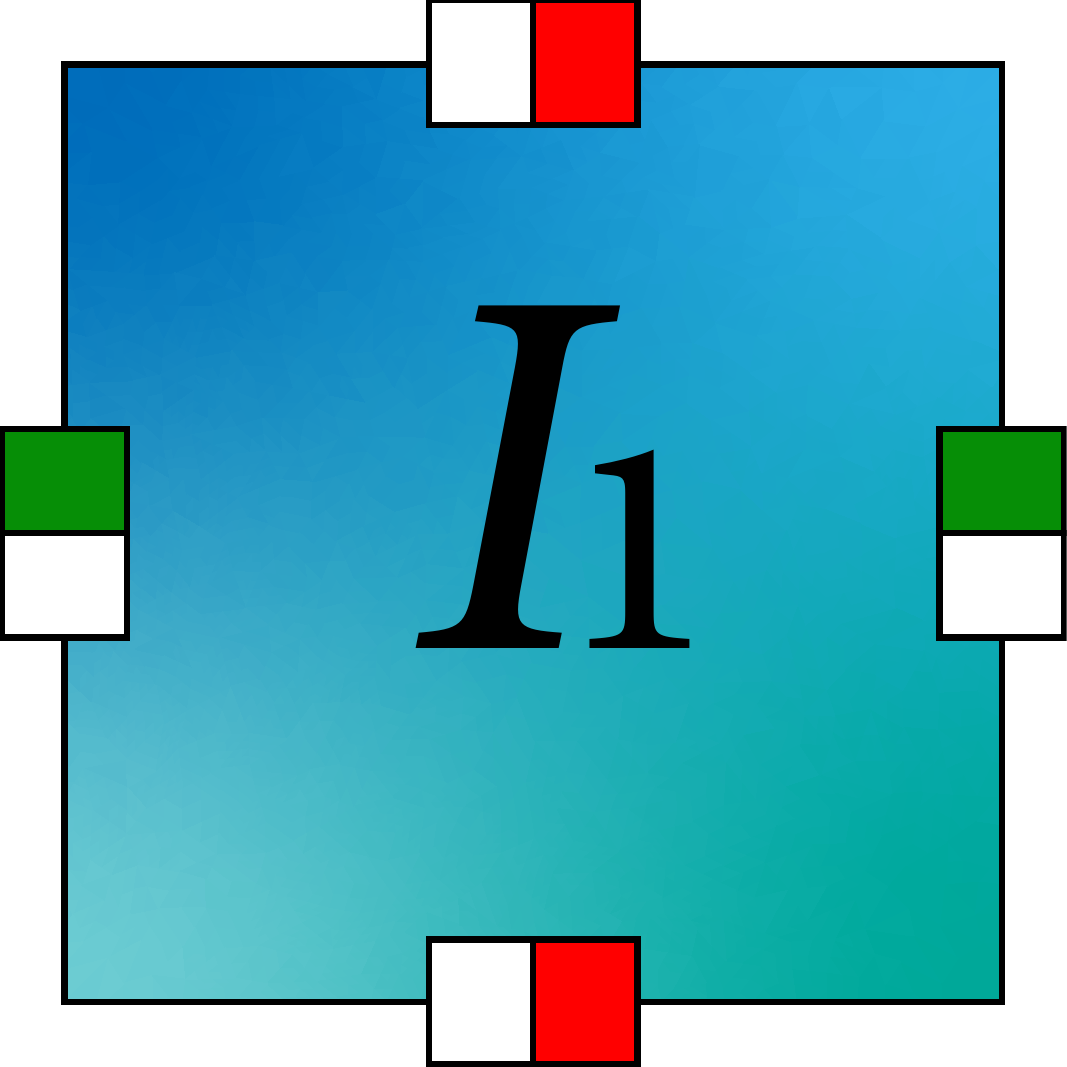} &
        \multicolumn{2}{c}{\includegraphics[height=\intro]{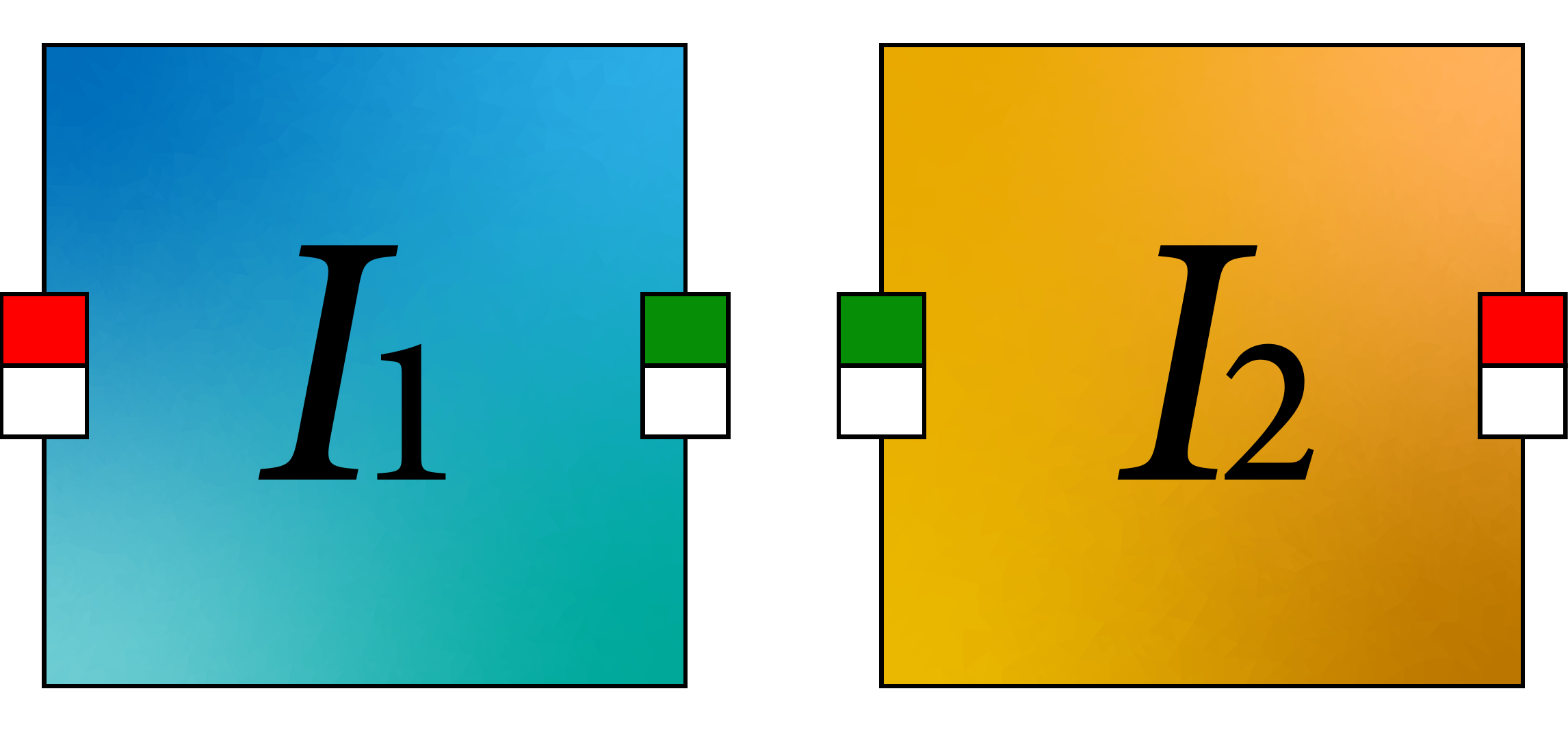}} &
        \multicolumn{2}{c}{\includegraphics[height=\intro]{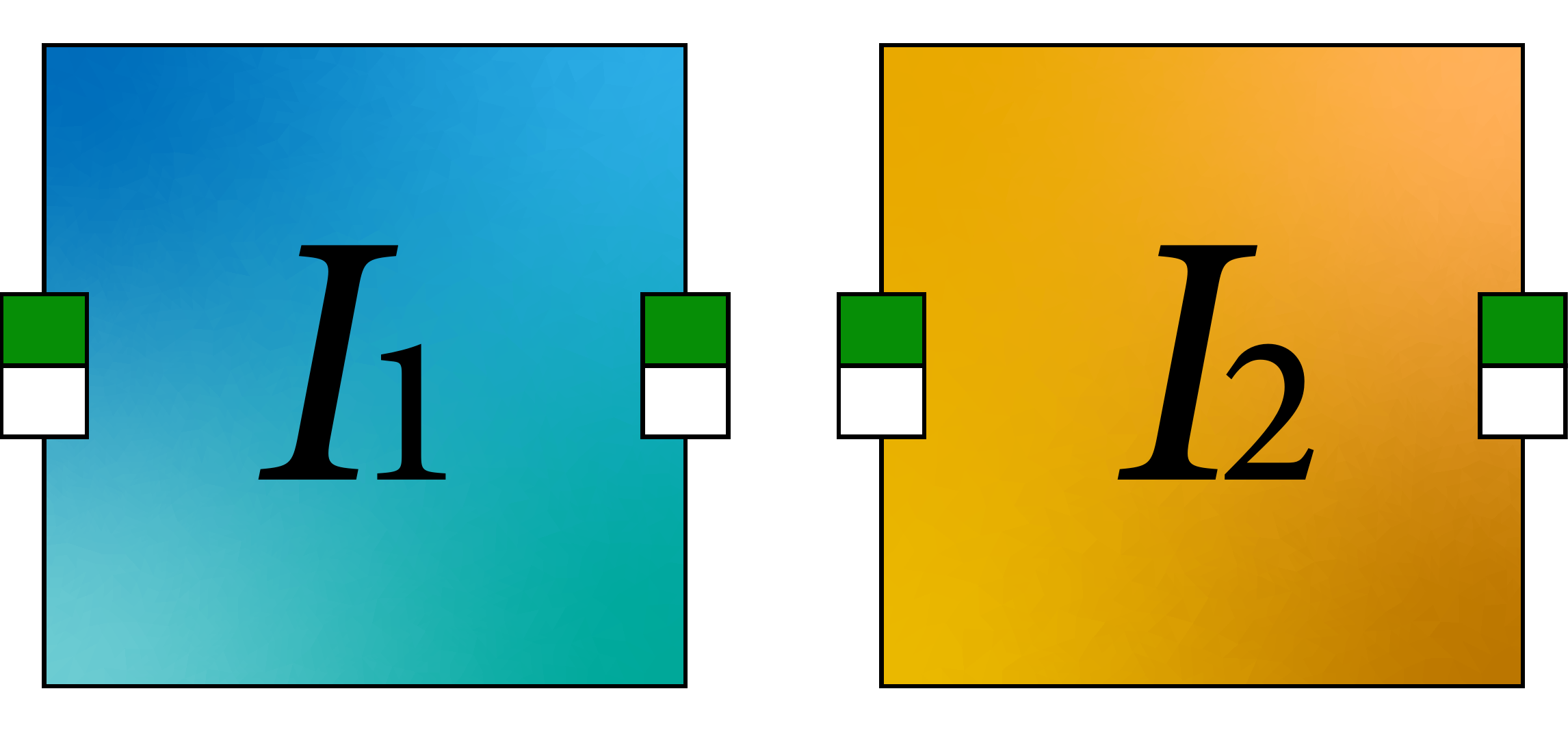}} \\
        
        \tiny{$\{\{I_{1|}\}, \{I_{|1}\}\}$} &
        \multicolumn{2}{c}{\tiny{$\{\{I_{1|}\}, \{I_{|2}\}\}$}} &
        \multicolumn{2}{c}{\tiny{$\{\{I_{1|}, I_{2|}\}, \{I_{|1}, I_{|2}\}\}$}} \\

        \tiny{$\{\{I_{\overline{1}}\}, \{I_{\underline{1}}\}\}$} &
        \multicolumn{2}{c}{\tiny{$\{\{I_{2|}\}, \{I_{|1}\}\}$}}
        &
        &
        \\
    \end{tabular}
    
    \caption{\textbf{Tiling scenarios.} Left: Self-tiling, where the image connects only to itself vertically and horizontally. Middle: One-to-one tiling on the X-axis, with each image connecting only to the other image. Right: Many-to-many tiling on the X-axis, where right sides of both images can connect to left sides of both images.
    Lower: constraints ($C_j$'s) for each tiling scenario. }
    \label{intro:example}
\end{figure}

%% file: sec/2_related.tex
\section{Related Work}
\label{sec:related}
We begin by discussing classical tiling techniques, highlighting the evolution of this field. We then explore recent advancements in image generation models, focusing on those applicable to image synthesis and editing. Next, we review tiling in popular applications. Finally, we examine current approaches to automated tiling using generative models, emphasizing the gaps our work aims to address.

\subsection{Classical Tiling Techniques}
\label{sec:related:classical}
Tiling, the seamless arrangement of patterns or images, has been a significant area of research in computer graphics and image processing. The introduction of Wang Tiles for image and texture generation allowed for non-periodic tiling of larger areas using squares with colored edges \cite{cohen2003wang}. This concept was further expanded with an alternative approach using colored corners \cite{Lagae2006AnAF} and the development of recursive Wang Tiles for real-time blue noise generation \cite{Kopf2006RecursiveWT}. A more theoretical approach demonstrated aperiodic tiling of the plane using a single prototile \cite{SOCOLAR20112207}. These classical approaches laid the foundation for efficient tiling techniques, addressing challenges in creating diverse, seamless patterns while balancing computational efficiency and visual quality.
\subsection{Image Generation and Editing Models}
\label{sec:related:image_models}
Recent years have seen significant advancements in image generation and editing, driven by deep learning models. While Generative Adversarial Networks (GANs) \cite{goodfellow2014generative} and Variational Autoencoders (VAEs) \cite{kingma2013auto} have shown promise, diffusion models \cite{sohl2015deep} have emerged as a particularly powerful technique for high-quality image generation.
Diffusion models operate by gradually adding noise to an image and then learning to reverse this process \cite{ho2020denoising}. This approach has led to improved sample quality and model robustness compared to earlier methods. Diffusion models have demonstrated impressive results in various applications, including text-to-image synthesis \cite{ramesh2022hierarchical}, image super-resolution \cite{saharia2022image}, and image editing \cite{meng2021sdedit,zabari2023diffusingcolorsimagecolorization,Avrahami_2022_CVPR,avrahami2023blendedlatent}. For a comprehensive overview of image synthesis and editing using diffusion models, we refer readers to \citet{huang2024diffusionmodelbasedimageediting} and \citet{cao2023comprehensive}.
We mostly use the Stable Diffusion 1.5 architecture \cite{rombach2022high}, as the generative backbone of our tiling method, but we also use other latent diffusion architectures \cite{podell2023sdxlimprovinglatentdiffusion,esser2024scalingrectifiedflowtransformers,zhang2023addingconditionalcontroltexttoimage}. This choice allows us to leverage the high-quality image generation capabilities of diffusion models while extending their functionality to support complex tiling scenarios.

\subsection{Related Applications of Tiling}
\label{sec:related:applications}

\subsubsection{Texture Synthesis}
There is vast research in creating tileable textures. Most recent approaches rely on input patches \cite{autoextract, zhou2024generatingnonstationarytexturesusing, zhou2023neural, Zhou_2018, Rodriguez_Pardo_2023, Sartor:2024:CAT, Rodriguez-Pardo_2024_CVPR}, multi-stage pipelines \cite{zhou2024generatingnonstationarytexturesusing, Fr_hst_ck_2019, Sartor:2024:CAT}, or specialized training \cite{Zhou_2018, Rodriguez_Pardo_2023, zhou2022tilegentileablecontrollablematerial, Rodriguez-Pardo_2024_CVPR}. Tiled Diffusion is unique in generating natural tileable textures directly from prompts using any latent diffusion model, not constrained to self-tiling scenarios.

\subsubsection{Infinite/Panorama Image Generation}
Unlike recent approaches for large-scale image generation \cite{bartal2023multidiffusionfusingdiffusionpaths, zhang2023diffcollageparallelgenerationlarge, lee2023syncdiffusion}, Tiled Diffusion generates a fixed set of tiles that can be seamlessly arranged in any order and quantity, offering a time- and space-efficient solution for infinite or panoramic imagery.

\subsection{Automated Tiling with Generative Models}
\label{sec:related:automated_tiling}
Seamless Tile Inpainting (STI) \cite{brick2face_seamless_tile} uses inpainting diffusion models to create seamless connections between swapped image quarters for self-tiling, or adjacent images for one-to-one connections. Asymmetric Tiling (AT) \cite{tjm35_asymmetric_tiling} modifies the Stable Diffusion architecture by replacing standard padding with circular padding, producing inherently tileable images but restricted to self-tiling scenarios with no rotations.

In contrast, our approach works directly on the latent representations during image generation. By creating all the different tiles simultaneously, we can share the necessary tiling information between the images being generated. This process makes our results inherently tileable across various domains, reducing inconsistencies and ensuring seamless connections.

We compare our method to STI and AT in Section \ref{sec:evaluation} as they are the most closely related methods to our work. We demonstrate that our approach offers greater flexibility in handling diverse tiling constraints across multiple images. Importantly, our method is unique in its ability to support complex many-to-many tiling scenarios outside the scope of texture synthesis, offering a significant advancement in the field of automated tileable image generation.

%% file: sec/3_method.tex
\section{Method}
\label{sec:method}
\input{figures/method/method_fig}
Our method generalizes the tiling problem to accommodate complex connections between multiple image sides. In this section, we first define the problem formally, then describe our approach to solving it using diffusion models.
\subsection{Problem Definition}
Given a set of desired output images $\mathcal{I} = \{I_1, I_2, ..., I_N\}$, where each image $I_i$ has four sides: right ($I_{i|}$), left ($I_{|i}$), top ($I_{\overline{i}}$), and bottom ($I_{\underline{i}}$), we define a set of tiling constraints $\mathcal{C} = \{C_1, C_2, ..., C_M\}$. Each constraint $C_j$ is represented as:
\begin{equation}
C_j = \{A_j, B_j\}
\end{equation}
Where $A_j$ and $B_j$ are sets of sides. These sets define the tileability of the output images. Specifically, any side from set $A_j$ is designed to seamlessly connect with any side from set $B_j$, and vice versa. Each side in $A_j$ and $B_j$ could be from any image in $\mathcal{I}$.

\subsection{Tiled Diffusion}
We use a diffusion model as the backbone of our method, operating on latent representations of images in a $H_{\text{latent}} \times W_{\text{latent}} \times D_{\text{latent}}$ latent space. The generation process begins with random Gaussian noise $\mathcal{N}(0, 1)$ and applies two key constraints at each diffusion step to ensure tiling coherence: a tiling constraint and a similarity constraint.

\subsubsection{Tiling Constraint}
\label{subsubsec:tiling}
The tiling constraint ensures global structure consistency across tiled images. This is achieved by copying a portion from a single latent representation in $B_j$ to pad the latents in $A_j$, and vice versa (copying from $A_j$ to $B_j$), for each constraint $C_j$. The way we choose the latent at each diffusion step is explained in \Cref{subsubsec:constraint_application}. The padding mechanism essentially enlarges the original latent representation to a larger size, which is then cropped back in pixel space after decoding the latent at the end of the diffusion process.
We use this approach because in the latent space of diffusion models, neighboring regions influence each other. By copying parts of the latent representation, we ensure that the generated images will have consistent content and style.

We define a context window of size $w$ for this operation, where $0 \leq w \leq H_{\text{latent}}/2$ or $W_{\text{latent}}/2$, depends on the tiling direction. The choice of $w$ affects the transition smoothness and variance between tiled images. Larger values of $w$ produce smoother transitions by copying more context, ensuring better global consistency. Smaller values allow for sharper transitions and more variance, preserving local details but potentially at the cost of some global coherence. \Cref{fig:context_window} illustrates the effect of different context window sizes.
This constraint is applied at each timestep of the diffusion process, ensuring that the tiling information is consistently maintained throughout the generation. By applying the constraint at every step, we guide the diffusion process to generate images that inherently satisfy the tiling requirements, rather than trying to enforce tiling as a post-processing step.

\subsubsection{Similarity Constraint}
\label{subsubsec:similarity}
The similarity constraint ensures seamless connections in complex many-to-many tiling scenarios, when multiple sides are involved in a single constraint ($|A_j| > 1$ or $|B_j| > 1$). For each set of sides involved in such a constraint, we copy the same latent representation over a small window that starts from the side itself and extends inward. We use a fixed window width of 5 in the latent space, which our experiments have shown to work effectively across all scenarios.

A key difference between the tiling constraint and the similarity constraint is their impact on the final result. The tiling constraint affects the result indirectly by providing context in regions that will be later cropped. In contrast, the similarity constraint directly affects the result by ensuring similarity in regions that are kept after cropping. This difference allows the tiling constraint to use much larger windows compared to the similarity constraint.

\subsubsection{Constraint Application Process}
\label{subsubsec:constraint_application}
For constraints involving different orientations (e.g., a top side and a left side), we rotate the latents accordingly.
For constraints involving multiple sides in $A_j$ or in $B_j$, we employ a round-robin mechanism. At each diffusion step, we cycle through the available sides, ensuring that over the course of the diffusion process, each side is exposed to all of its potential matches.
This approach allows the model to balance the influences of different constraints, resulting in coherently tiled outputs.

After applying all the constraints throughout the diffusion process, we decode the latent representations and crop the results to the original image dimensions, producing seamlessly tiled images that satisfy all predefined constraints. 
We provide visual ablation of both constraints in the supplementary material.

Figure \ref{fig:method} illustrates our Tiled Diffusion method with a many-to-many example, where the constraint $C_1 = \{\{I_{1|}, I_{2|}\}, \{I_{|1}, I_{|2}\}\}$ allows right sides of images 1 and 2 to connect seamlessly to left sides of both images. The figure shows input constraints, application of both constraint types, resulting images after decoding and cropping, and example tiling arrangements.

\subsection{Image-to-Image (Img2Img) Generation}
While the method described above focuses on text-to-image generation, our system also supports image-to-image (img2img) generation. In this setting, we start with an input image rather than random noise.
We begin by encoding the input image $\mathbf{x}$ using the Variational AutoEncoder (VAE) to obtain an initial latent representation $\mathbf{z}_0$:
\begin{equation}
\mathbf{z}_0 = \text{Encoder}(\mathbf{x})
\end{equation}
To start the diffusion process from a noisy latent $\mathbf{z}_t$, we add noise to $\mathbf{z}_0$ according to the diffusion schedule:
\begin{equation}
\mathbf{z}_t = \sqrt{\bar{\alpha}_t}\mathbf{z}_0 + \sqrt{1 - \bar{\alpha}_t}\boldsymbol{\epsilon}, \quad \boldsymbol{\epsilon} \sim \mathcal{N}(0, \mathbf{I})
\end{equation}
where $t$ is chosen to be sufficiently large, and $\bar{\alpha}_t$ is the cumulative product of noise schedule $\alpha_t$. The diffusion process then proceeds from $\mathbf{z}_t$, applying the same tiling and similarity constraints as in the text-to-image case.
For a more detailed explanation of img2img generation using diffusion models, we refer the reader to \citet{rombach2022high}.

In this paper, we primarily demonstrate our results on Stable Diffusion 1.5 and 2.0 \cite{rombach2022high}. However, we also showcase the applicability of our method to Stable Diffusion XL~\cite{podell2023sdxlimprovinglatentdiffusion}, Stable Diffusion 3~\cite{esser2024scalingrectifiedflowtransformers}, and ControlNet~\cite{zhang2023addingconditionalcontroltexttoimage} in the supplementary material.

\input{figures/opening/opening2}

%% file: figures/method/method_fig.tex
\newlength{\method}
\setlength{\method}{10cm}

\begin{figure*}[t]
    \centering
    \includegraphics[width=\textwidth]{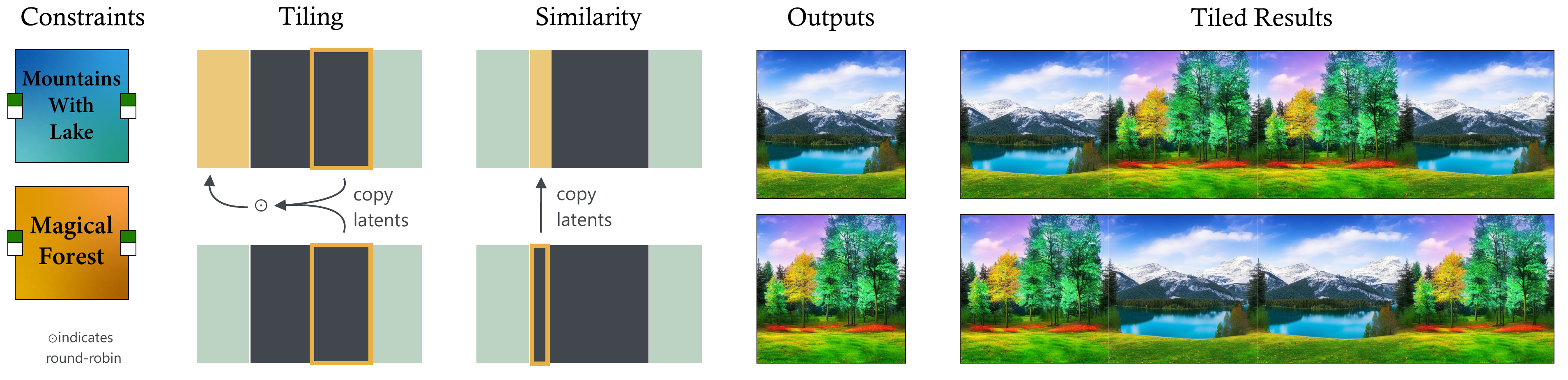}
\caption{
\textbf{Method overview and results.}
Our method uses two key constraints in latent space: tiling constraints for global consistency and similarity constraints for seamless connections in complex scenarios.
Left: Input constraints for many-to-many tiling on the X-axis between two images.
Second column: Tiling constraint applied to the left side of the first image, using round-robin context selection.
Third column: Similarity constraint propagated from the second image to the first.
Fourth column: Output images after decoding and cropping.
Right: Two example arrangements demonstrating many-to-many tiling scenarios.
}
    \label{fig:method}
\end{figure*}

%% file: figures/opening/opening2.tex
\newlength{\www}
\setlength{\www}{1.3cm}

\newlength{\wwww}
\setlength{\wwww}{0.99\columnwidth}

\begin{figure}[t]

        \centering
        \setlength{\tabcolsep}{0cm} %
        \begin{tabular}{c}
            $w=32$ \\
            \includegraphics[width=\wwww, frame]{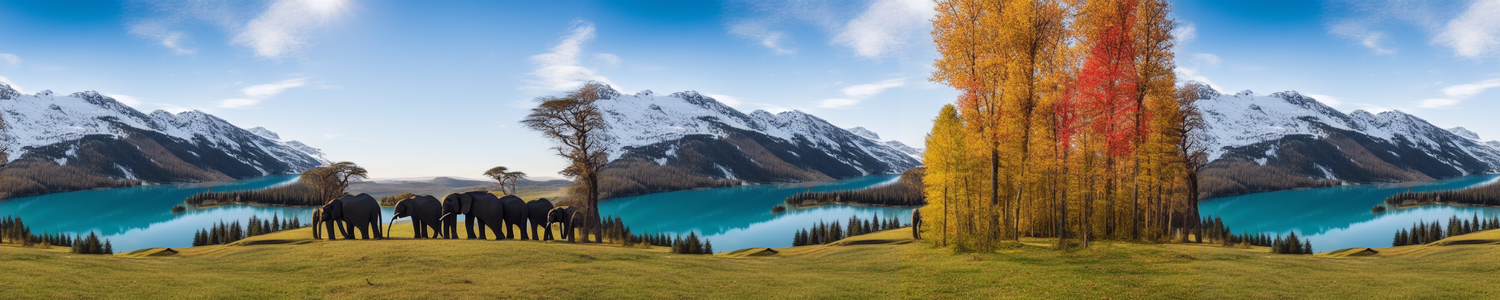} \\ 
            $w=20$\\
            \includegraphics[width=\wwww, frame]{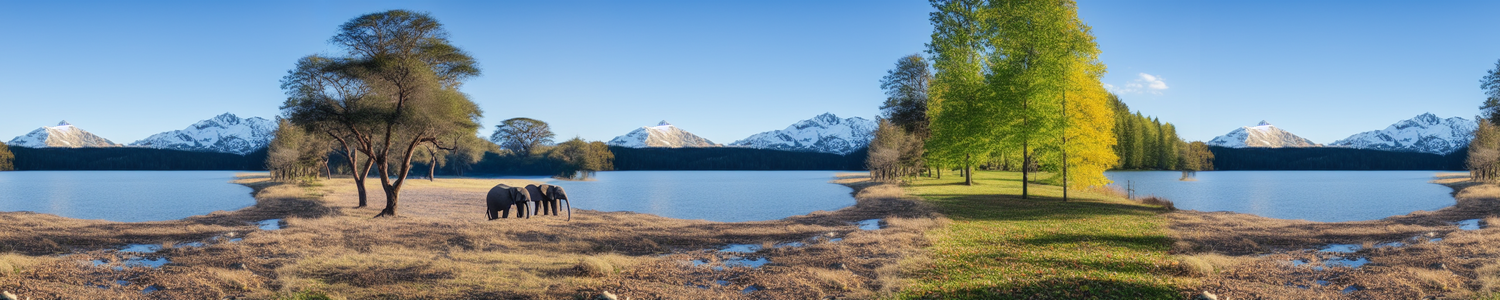} \\ 
            $w=8$\\
            \includegraphics[width=\wwww, frame]{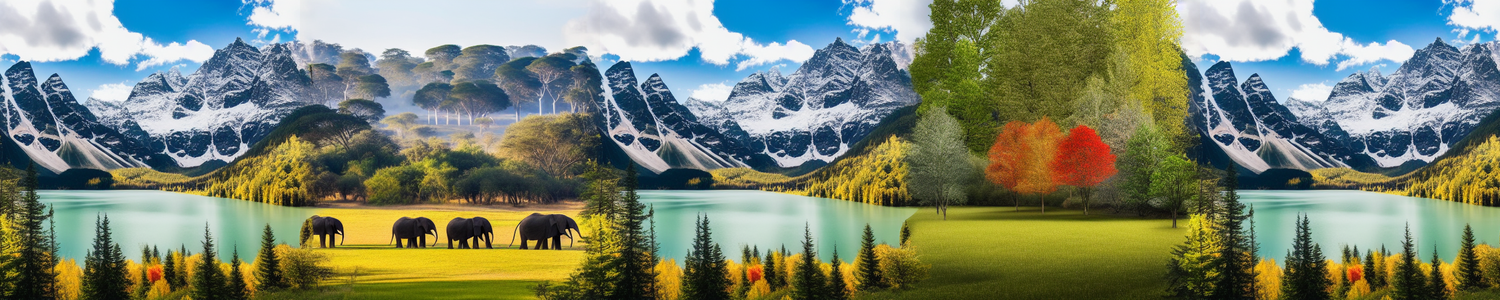} \\ 
        \end{tabular}
    \caption{
Illustration of the impact of different context window sizes ($w$) on tiling results. The figure displays panoramic views created by horizontally stacking the results for large (top), medium (middle), and small (bottom) $w$. We observe that with a large $w$, the transitions between tiles are smoother and more gradual, resulting in a more coherent overall image with less transition variations. As $w$ decreases, we see sharper transitions variations, but potentially at the cost of global coherence.
}

    \label{fig:context_window}
\end{figure}

%% file: sec/4_evaluation.tex
\section{Evaluation}
\label{sec:evaluation}

Our evaluation includes qualitative and quantitative analyses compared to STI \cite{brick2face_seamless_tile} and AT \cite{tjm35_asymmetric_tiling}. We visually compare image quality and tiling capabilities, showcasing complex tiling arrangements. The numerical metrics assess our method's performance in the different tiling scenarios, with comparisons to STI and AT where applicable.
\subsection{Qualitative Evaluation}
All methods support self-tiling. The one-to-one tiling is supported by our method and STI. The many-to-many tiling scenarios are exclusively supported by our method.
\Cref{fig:qualitative:self} compares self-tiling results of Tiled Diffusion, STI, and AT. In the landscape examples (rows 1-2), our method and AT maintain logical continuity across tiles, with coherent alignment of elements like trees and clouds. STI, however, shows misaligned elements and artifacts at connection points. For the texture example (row 3), all methods perform well, as it lacks complex high-level structure.
\Cref{fig:qualitative:one-to-one} demonstrates one-to-one tiling scenarios, comparing Tiled Diffusion and STI. These examples highlight our method's ability to generate coherent macro details between the two images. Our results show logically connected images, while the STI method struggles to maintain consistency across tile boundaries. This is because STI generates each image independantly, while our method shares information accross the images during synthesis.
\Cref{fig:qualitative:many-to-many} showcases many-to-many tiling scenarios, which are uniquely supported by our Tiled Diffusion method. These examples highlight our method's ability to handle complex tiling arrangements while maintaining visual coherence.
Additional qualitative results and comparisons can be found in the supplementary document.

\subsection{Quantitative Evaluation}
\label{sec:evaluation:quantitative}
To evaluate our method, we used a random sample of 1,000 captions from the LAION 400M dataset \cite{schuhmann2021laion400mopendatasetclipfiltered}. We assess the performance of our method across three scenarios: self-tiling, one-to-one tiling, and many-to-many tiling. Our evaluation aims to measure the quality of tiling, the fidelity to the input prompts, and the overall image quality.
We employ four evaluation metrics to analyze these aspects:
\paragraph{Tiling Score (TS).} We propose TS to measure tiling quality across various image domains. While existing methods for assessing tiling quality, like TexTile \cite{Rodriguez-Pardo_2024_CVPR}, focus on textures and are unable to generalize to other domains, our method is applicable to a broader range of images (examples in supplementary material). TS is calculated as the maximum of three separate measurements between two tiled images to capture potential discontinuities beyond the immediate connection area. Each measurement is defined as the mean absolute difference of pixel values at specific locations. Lower values indicate better tiling (range: 0.0 to 1.0).
For a pair of images $I_1$ and $I_2$ tiled along an axis of length $h$:
\begin{equation}
TS_{pair}(I_1, I_2) = \frac{1}{h} \sum_{y=1}^{h} |I_1(x_1, y) - I_2(x_2, y)|
\end{equation}
\noindent
where $x_1$ and $x_2$ are the column indices of the adjacent edges. This equation is applied three times: once at the connection area, and twice at fixed offsets to the left and right (or top and bottom) of the connection. The overall TS is the average of these three measurements.
This approach helps identify discontinuities that might occur at the boundaries of inpainted regions.
To validate TS, we designed a procedure to test its effectiveness on well-connected images. We applied TS to 1,000 images generated from LAION captions using vanilla text-to-image generation (VT2I), as well as to versions of these images with their left and right halves, top and bottom halves swapped (swapped VT2I). The idea behind this procedure is that the swapped images should have perfect connections at the swapped edge, resulting in low TS scores. As expected, the swapped VT2I images achieved significantly lower TS scores compared to the original VT2I images, confirming the validity of our metric.

\paragraph{Additional Metrics.} To measure image-text similarity (CLIP-Score), we use the cosine similarity between the CLIP encodings~\cite{clip} of the image and the text. For perceptual similarity, we use LPIPS~\cite{zhang2018unreasonable}, which assesses the diversity of generated images compared to the VT2I images, with higher values indicating more diverse outputs. For overall image quality and realism, we use FID~\cite{heusel2017gans} between the generated images and VT2I.
We also include a baseline evaluation for VT2I and Swapped VT2I, where we additionally calculate the same metrics for both. This baseline allows us to assess how our method performs relative to standard text-to-image generation.
\Cref{tab:statistical_analysis} shows that our method performs well across all metrics compared to other approaches. For this analysis, we evaluated tileability on the X-axis using the sampled prompts. For self-tiling, we created 1,000 constraints (one per prompt). For one-to-one tiling, we randomly formed 500 pairs of prompts, resulting in 500 constraints. In many-to-many scenarios, each constraint $C_j$ involved sets $A_j$ and $B_j$ with 2 sides each, using 4 prompts per constraint, totaling 250 constraints.
Our Tiled Diffusion method performs consistently across all scenarios. The table also demonstrates the impact of removing key components of our method. Without the tiling constraint (TC), we achieve results similar to standard image generation --- high quality but poor tiling. Without the similarity constraint (SC), performance remains stable except in many-to-many scenarios, where tiling quality decreases. This highlights the importance of both constraints in our method.
\Cref{fig:scalability} illustrates how each metric (TS, CLIP-Score, LPIPS, FID) scales with increasing complexity in many-to-many tiling scenarios. Here, $n$ represents the number of sides within sets $A_j$ and $B_j$ for each constraint $C_j$, with $n=1$ corresponding to one-to-one tiling. As $n$ increases, the results remain nearly constant, demonstrating that our method scales effectively to larger sets of constraints while maintaining tileability and quality.

Additionally, we provide texture synthesis analysis in the supplemental material where we compare our method to other methods in example-based texture synthesis, and evaluate our method in generating textures directly from prompts.\input{figures/quantitative/scalability-new}

\input{tables/quantitative}

%% file: figures/quantitative/scalability-new.tex
\newlength{\tablewidth}
\setlength{\tablewidth}{4.6cm}

\begin{figure}[t]
    \centering
        \centering
        \setlength{\tabcolsep}{5pt} %
        \begin{tabular}{ c c }
            \includegraphics[width=0.45\columnwidth]{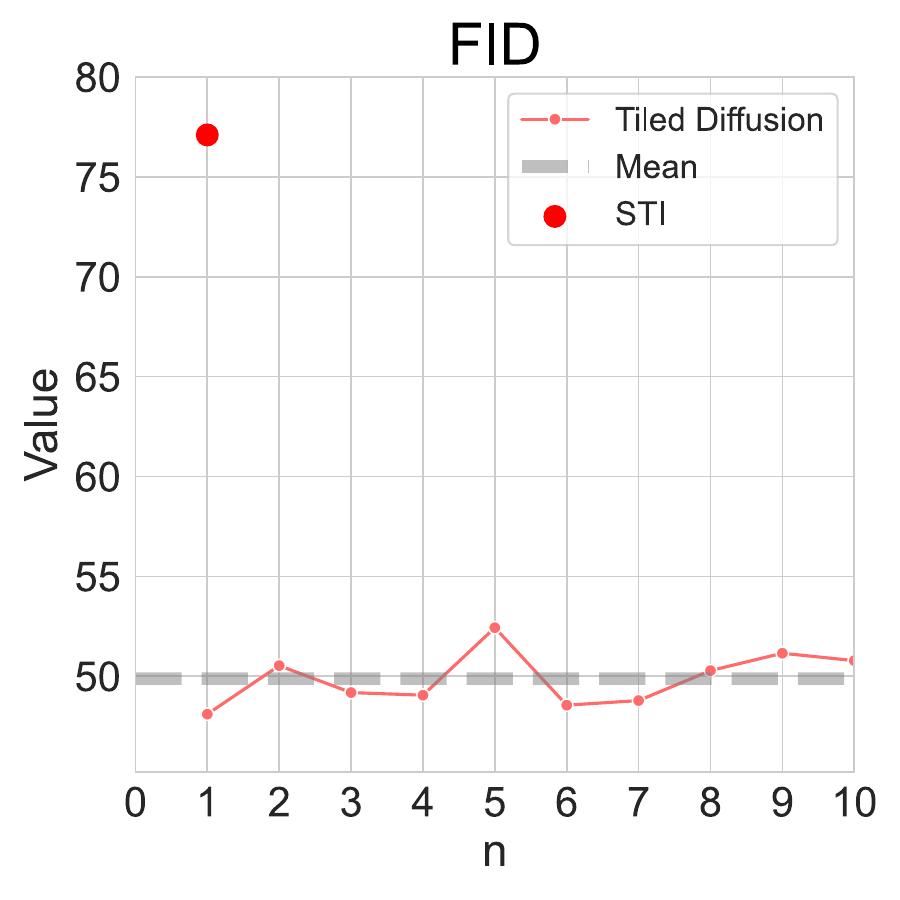} &
            \includegraphics[width=0.45\columnwidth]{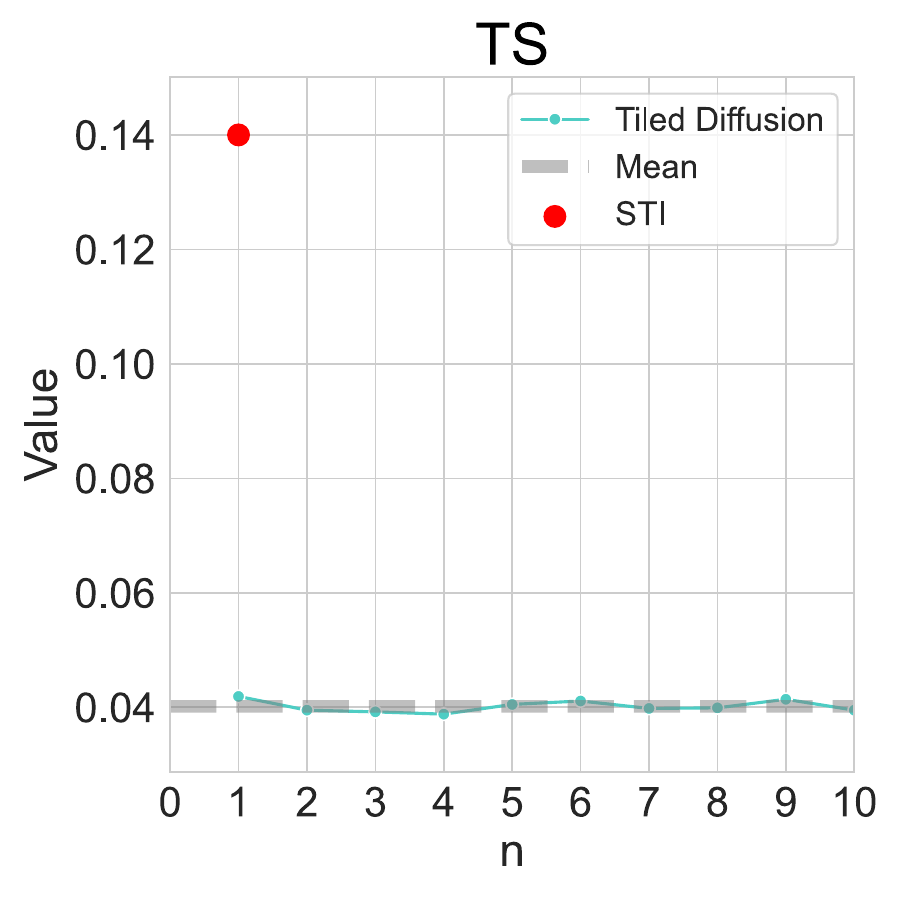} \\
            \includegraphics[width=0.45\columnwidth]{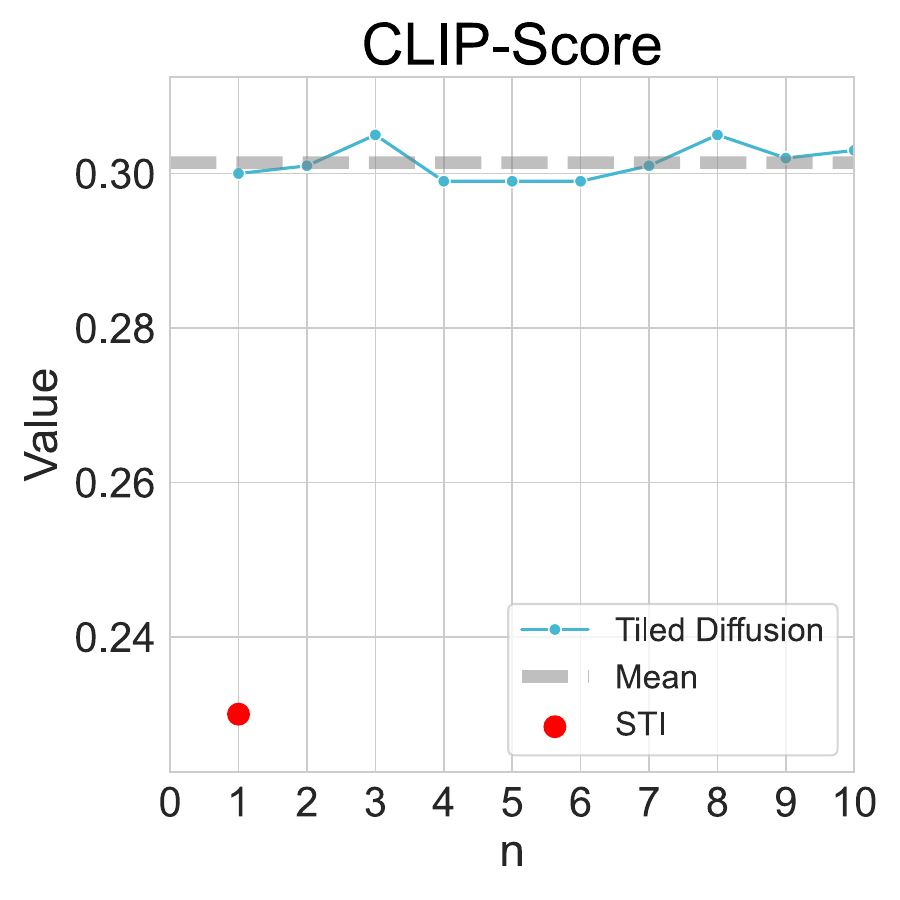} &
            \includegraphics[width=0.45\columnwidth]{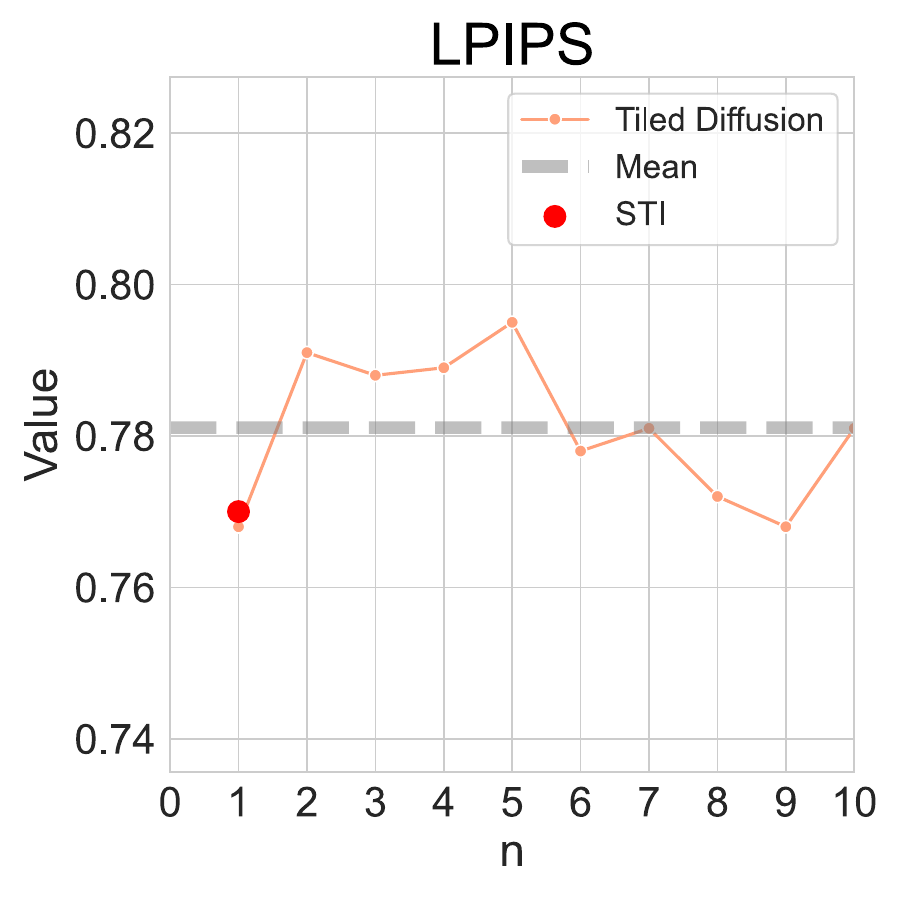} \\
        \end{tabular}
\caption{Quantitative analysis of our metrics as we increase the number of sides in $A_j$ and $B_j$ for each constraint $C_j$ in many-to-many tiling scenarios ($n=|A_j|=|B_j|$). The graphs show relatively constant values across different side counts, demonstrating our method's ability to scale effectively. For $n = 1$ (one-to-one) we also include the STI method (red dot).}    \label{fig:scalability}
\end{figure}

%% file: tables/quantitative.tex
\begin{table*}[t]
    \renewcommand{\arraystretch}{1.2}
    \centering

    \begin{adjustbox}{width=\textwidth}
        \begin{tabular}{l cccc cccc cccc}
            \toprule
            \multirow{2}{*}{\textbf{Method}}
            & \multicolumn{4}{c}{\textbf{Self-Tiling}} & \multicolumn{4}{c}{\textbf{One-To-One}} & \multicolumn{4}{c}{\textbf{Many-To-Many}} \\
            \cmidrule(lr){2-5} \cmidrule(lr){6-9} \cmidrule(lr){10-13} 
            & FID $\downarrow$ &  TS $\downarrow$& CLIP-Score $\uparrow$& LPIPS $\downarrow$& FID $\downarrow$& TS $\downarrow$& CLIP-Score $\uparrow$& LPIPS $\downarrow$& FID $\downarrow$& TS $\downarrow$& CLIP-Score $\uparrow$& LPIPS $\downarrow$\\
            \cmidrule(lr){2-2} \cmidrule(lr){3-3} \cmidrule(lr){4-4} \cmidrule(lr){5-5} 
            \cmidrule(lr){6-6} \cmidrule(lr){7-7} \cmidrule(lr){8-8} \cmidrule(lr){9-9} 
            \cmidrule(lr){10-10} \cmidrule(lr){11-11} \cmidrule(lr){12-12} \cmidrule(lr){13-13} 
            VT2I & \xmark & 0.29 & 0.31 & \xmark & \xmark & 0.29 & 0.31 & \xmark & \xmark & 0.29 & 0.31 & \xmark  \\
            Swapped VT2I & \xmark & 0.03 & 0.14 & \xmark & \xmark & 0.03 & 0.14 & \xmark & \xmark & 0.03 & 0.14 & \xmark  \\
            \midrule
            AT & 49.2 & 0.03 & 0.29 & 0.79 & \xmark & \xmark & \xmark & \xmark & \xmark & \xmark & \xmark & \xmark \\
            STI & 59.2 & 0.03 & \textbf{0.31} & 0.77 & 77.1 & 0.14 & 0.23 & 0.77 & \xmark & \xmark & \xmark & \xmark \\
            Tiled Diffusion & \textbf{47.9} & 0.03 & 0.30 & \textbf{0.76} & \textbf{48.1} & \textbf{0.04} & \textbf{0.30} & \textbf{0.76} & \textbf{50.5} & \textbf{0.03} & \textbf{0.30} & \textbf{0.79} \\
            \midrule
            \textsmaller[1]{Tiled Diffusion w/o TC} & 48.1 & 0.28 & 0.30 & 0.77 & 48.4 & 0.31 & 0.30 & 0.77 & 50.9 & 0.30 & 0.30 & 0.79 \\
            \textsmaller[1]{Tiled Diffusion w/o SC} & 47.9 & 0.03 & 0.30 & 0.76 & 48.1 & 0.04 & 0.30 & 0.76 & 50.5 & 0.12 & 0.30 & 0.79 \\
            \bottomrule
        \end{tabular}
    \end{adjustbox}
    \caption{\textbf{Quantitative Analysis.} Comparison across self-tiling, one-to-one, and many-to-many scenarios. Top: pseudo ground-truth . VT2I (1000 Stable Diffusion-generated images) and Swapped VT2I (with halves swapped on X and Y axes) are non-tiled and tiled-by-definition images, respectively. Middle: comparisons. Our Tiled Diffusion method outperforms AT and STI in most comparable scenarios (self-tiling and one-to-one), and uniquely supports many-to-many tiling. Bottom: Ablations. 
    Without the tiling constraint (TC), images are not tiled (high TS, similar to VT2I); without the similarity constraint (SC), tiling performance decreases in many-to-many scenarios (higher TS).}
    \label{tab:statistical_analysis}
\end{table*}

%% file: sec/5_applications.tex
\input{figures/qualitative/qualitative}
\input{figures/qualitative/many-to-many}
\section{Applications}
\label{sec:applications}

Our Tiled Diffusion method enables various applications in image generation and processing. We highlight three: Seamlessly Tiling Existing Images, Tileable Texture Generation, and 360° Synthesis, as illustrated in \Cref{teaser}. Additional examples and analysis are provided in the supplementary material.

\subsection{Seamlessly Tiling Existing Images}
Many digital images are not inherently tileable, limiting their use in repetitive patterns. Our method, combined with Differential Diffusion \cite{levin2023differential}, transforms non-tileable images into seamlessly tileable versions while preserving most of the original content.
\subsection{Tileable Texture Generation}
Our Tiled Diffusion method, applied to the SDXL model \cite{podell2023sdxlimprovinglatentdiffusion}, generates seamlessly tileable textures. This enables creation of diverse, coherent texture sets and mix-and-match tiles (\Cref{fig:qualitative:many-to-many} top row), enhancing realism in 3D modeling, game development, and digital art.
\subsection{360° Synthesis}
Our Tiled Diffusion method creates wraparound panoramic images by generating seamlessly connected edges. This technique produces continuous views for wide-angle landscapes or architectural panoramas. 

%% file: figures/qualitative/qualitative.tex
\newlength{\selfwidthc}
\setlength{\selfwidthc}{1.8cm}

\newlength{\selfwidthr}
\setlength{\selfwidthr}{3.8cm}

\newlength{\selfheight}
\setlength{\selfheight}{1.63cm}

\begin{figure*}[ht]
    
    \centering
        \centering
        \setlength{\tabcolsep}{3pt} %
        \begin{tabular}{cccc}
             & Tiled Diffusion & STI & AT \\[0.1cm]
            
            \includegraphics[height=\selfheight]{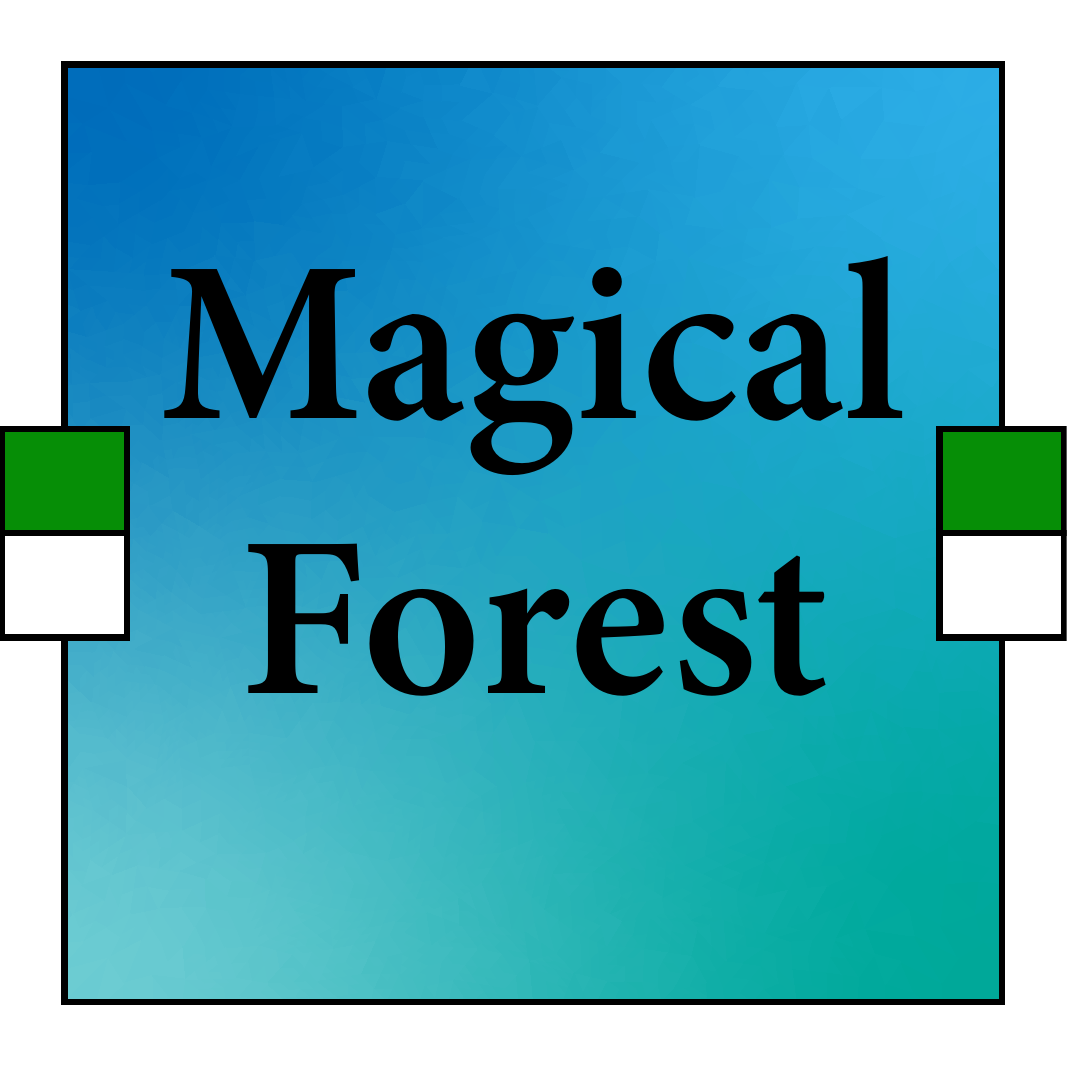} &
            \includegraphics[height=\selfheight,frame]{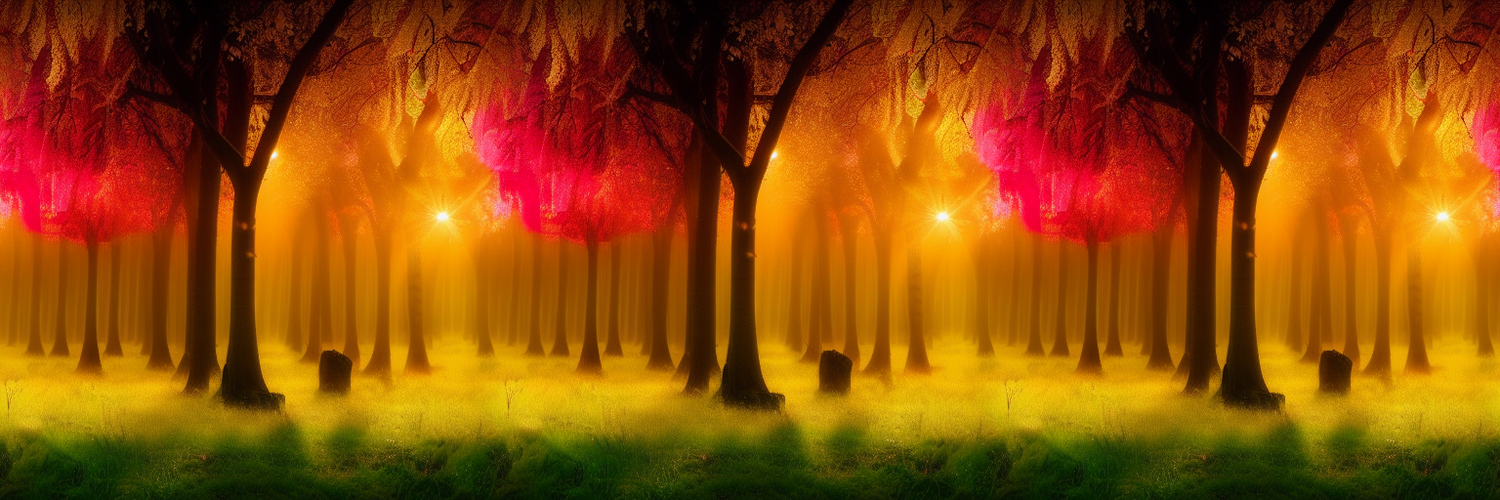} &
            \includegraphics[height=\selfheight,frame]{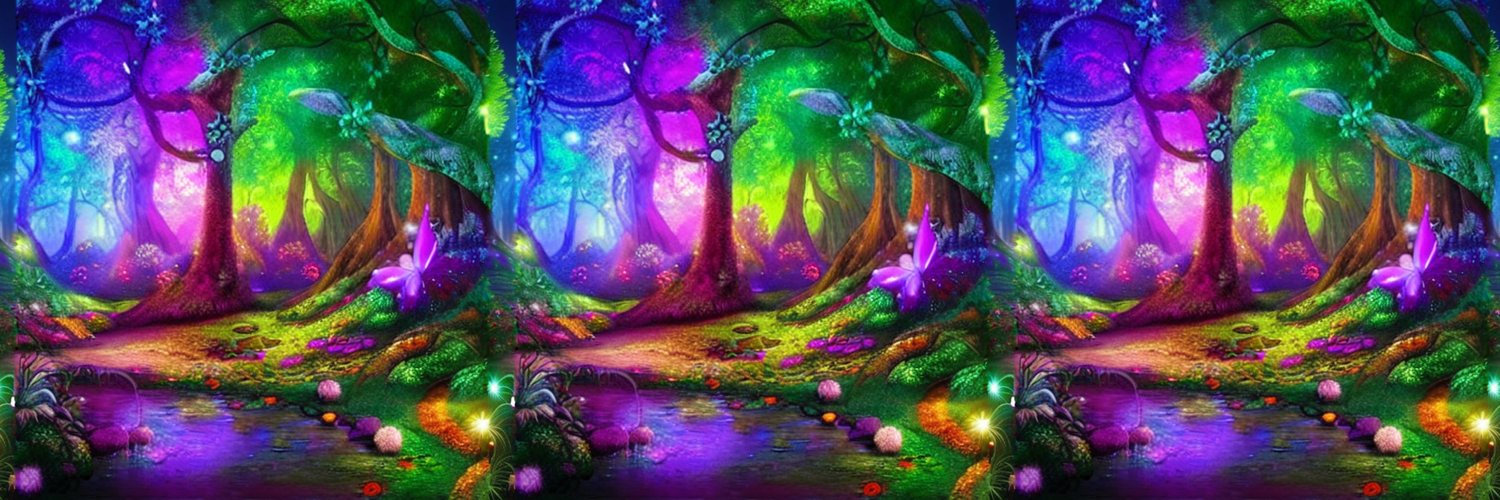} &
            \includegraphics[height=\selfheight,frame]{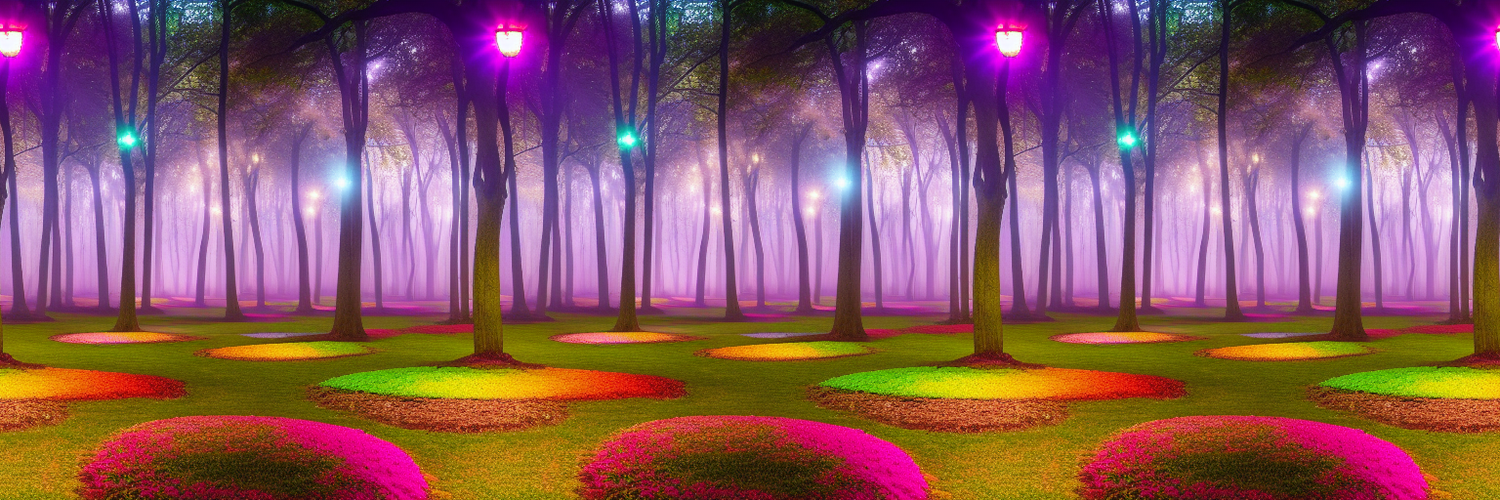} \\[0.1cm]

            \includegraphics[height=\selfheight]{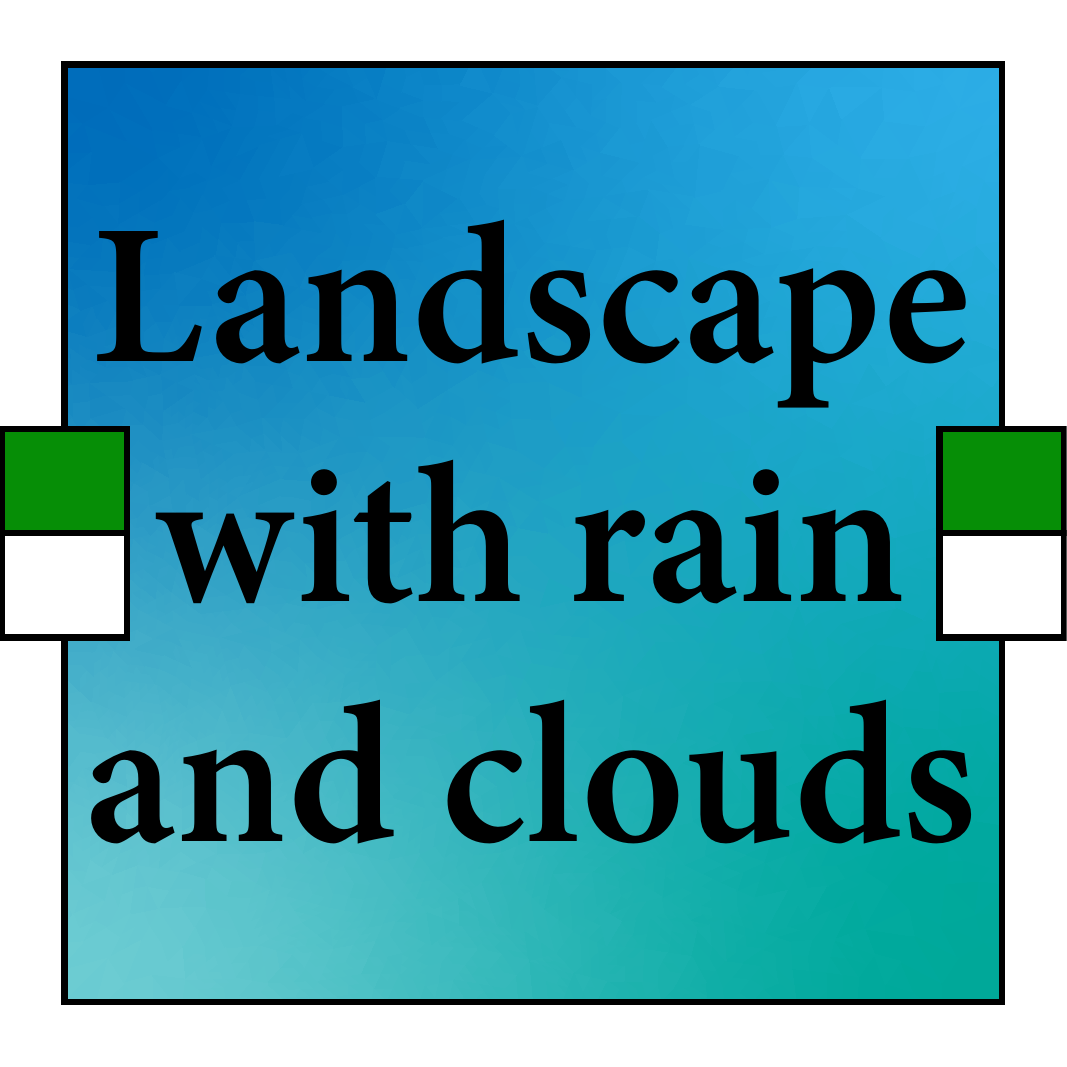} &
            \includegraphics[height=\selfheight,frame]{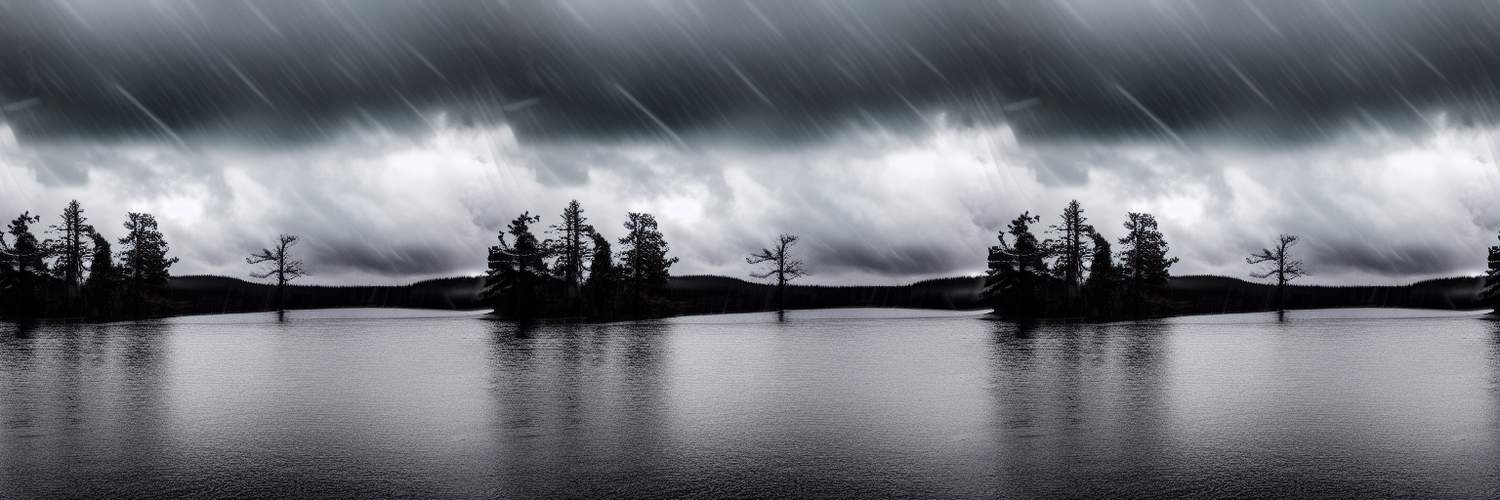} &
            \includegraphics[height=\selfheight,frame]{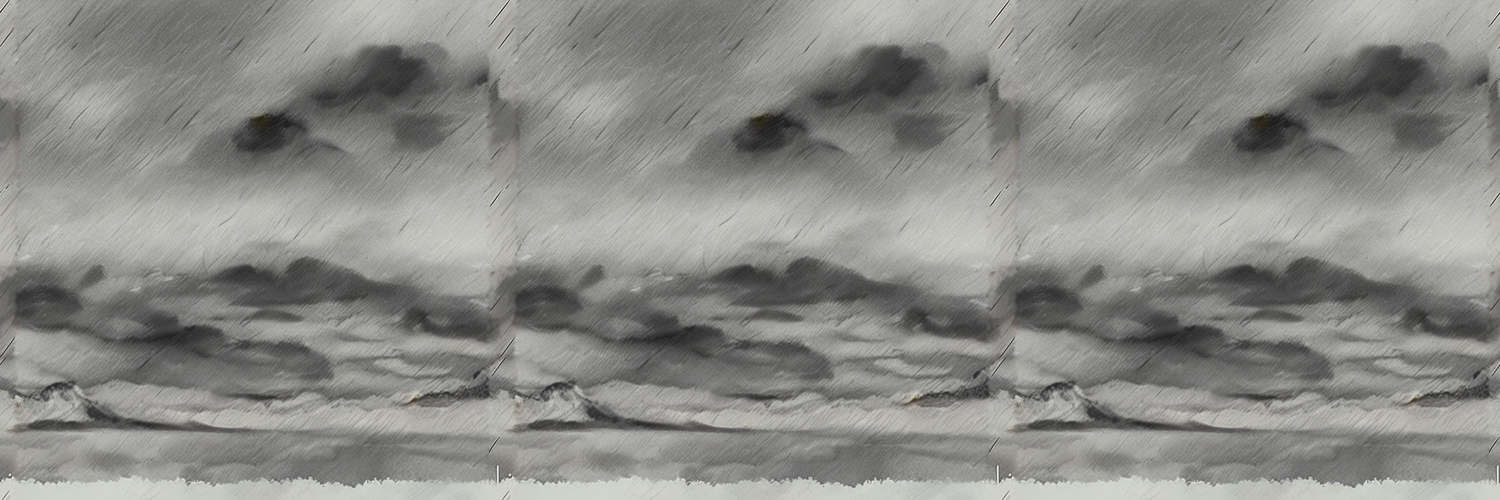} &
            \includegraphics[height=\selfheight,frame]{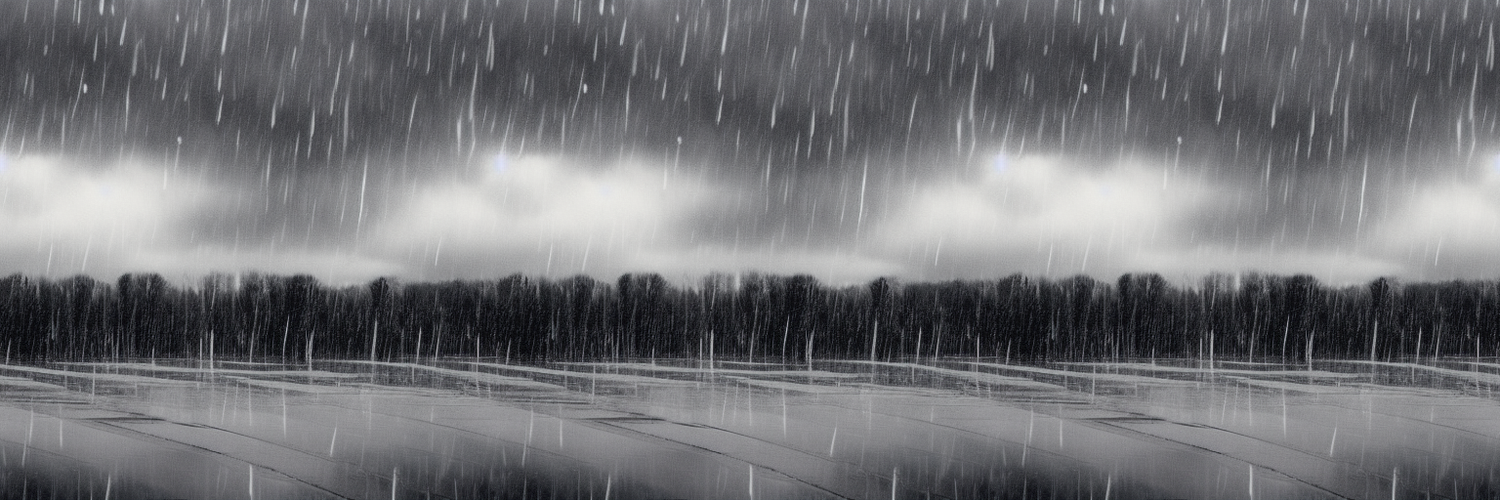} \\[0.1cm]

            \includegraphics[height=\selfheight]{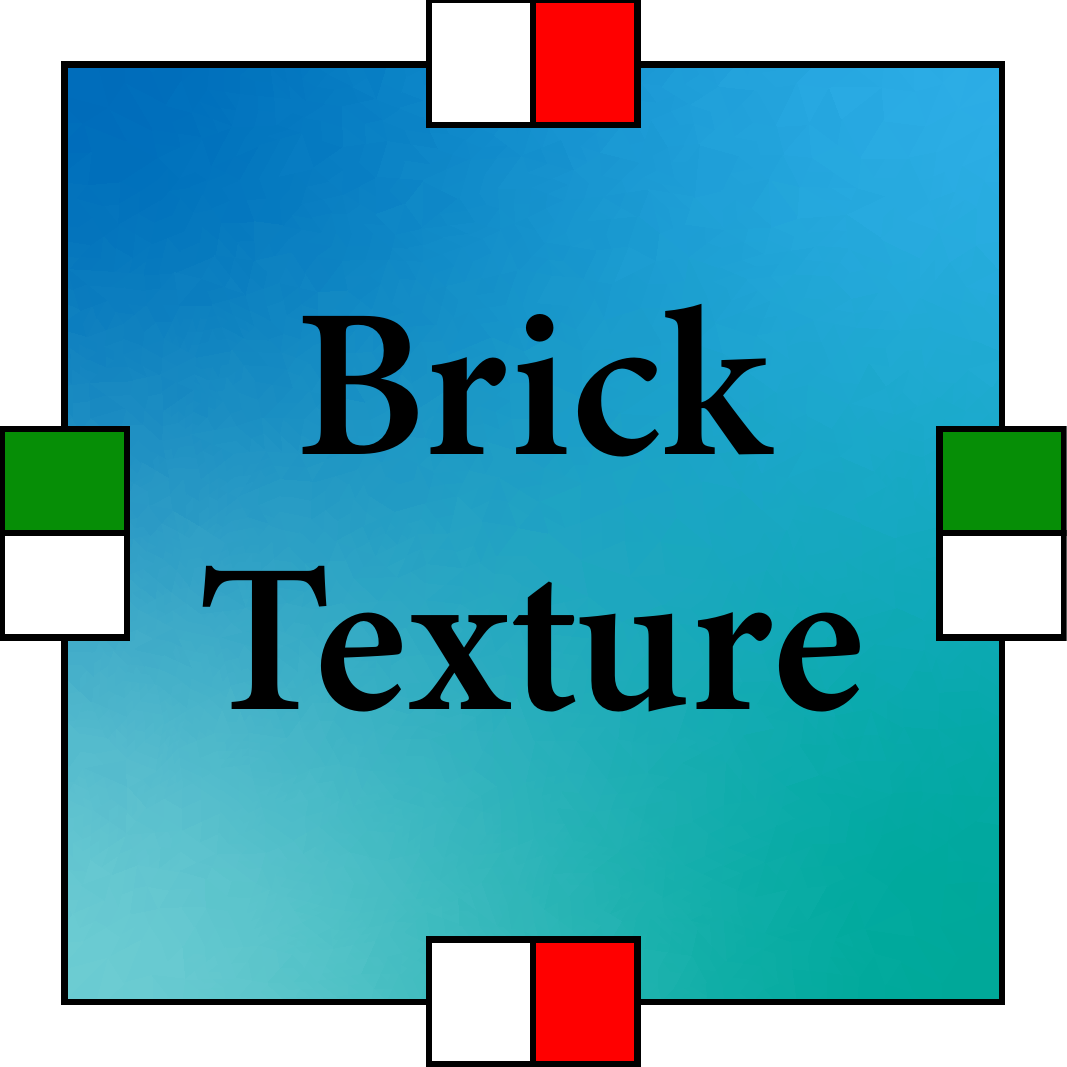} &
            \includegraphics[height=2\selfheight,frame]{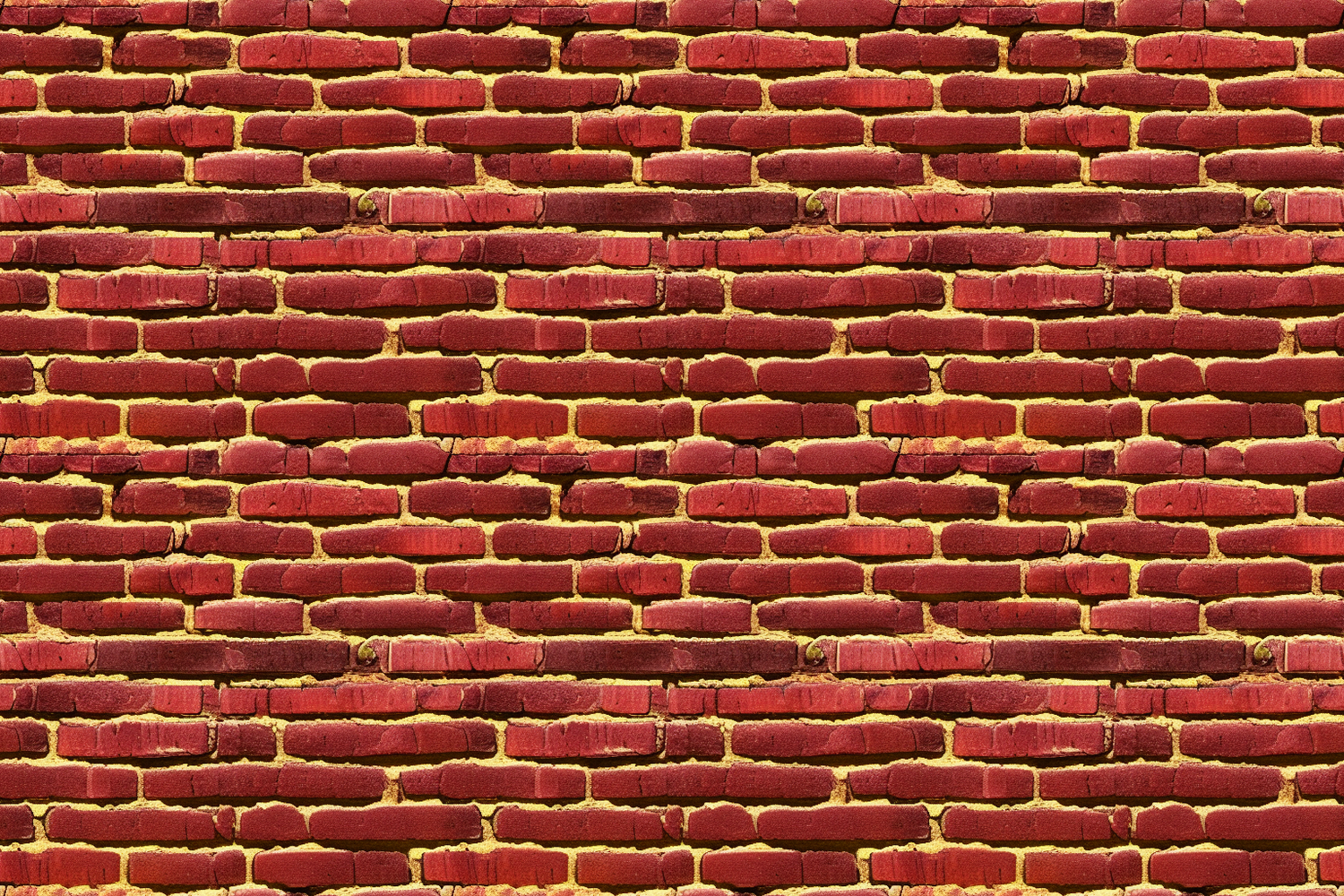} &
            \includegraphics[height=2\selfheight,frame]{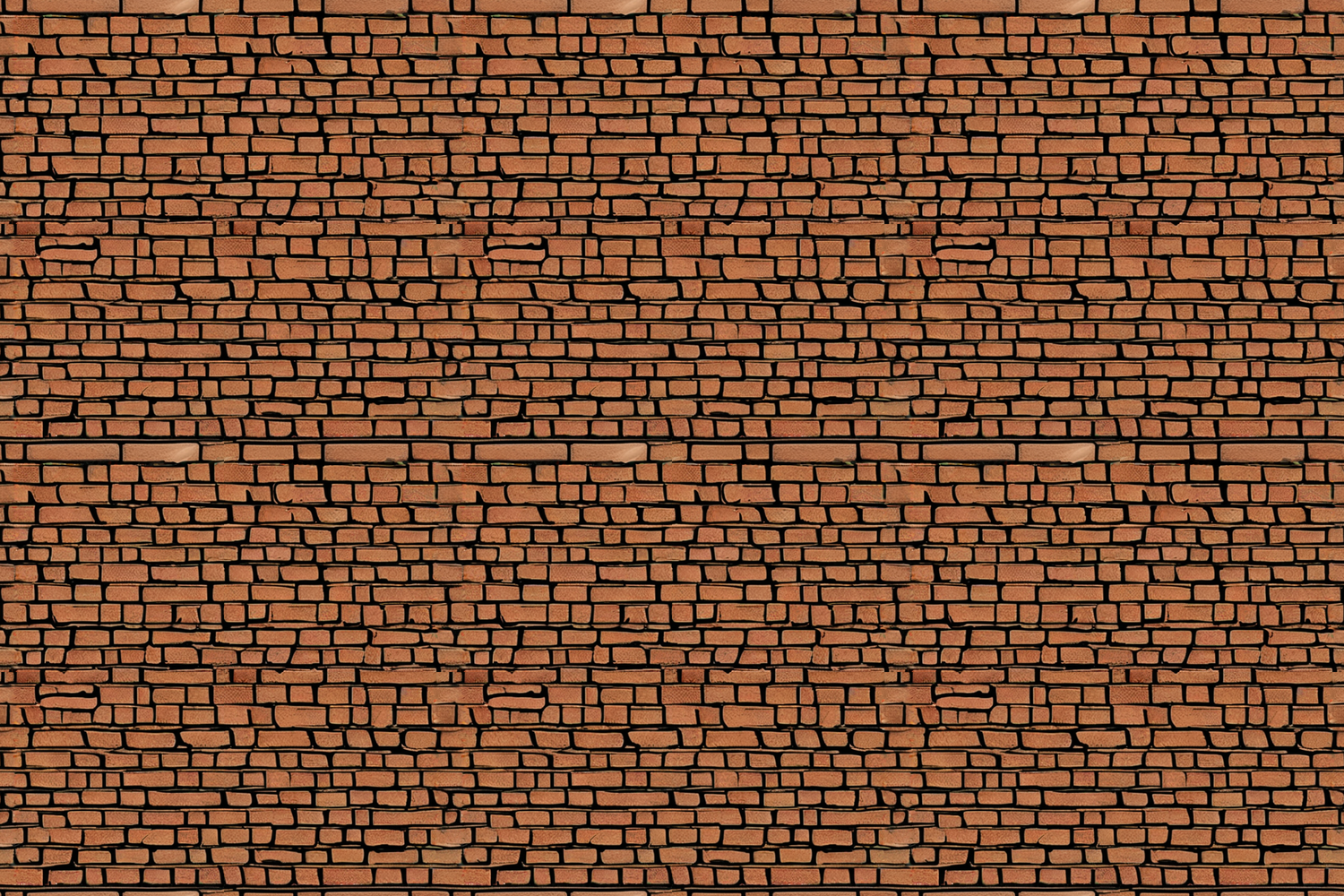} &
            \includegraphics[height=2\selfheight,frame]{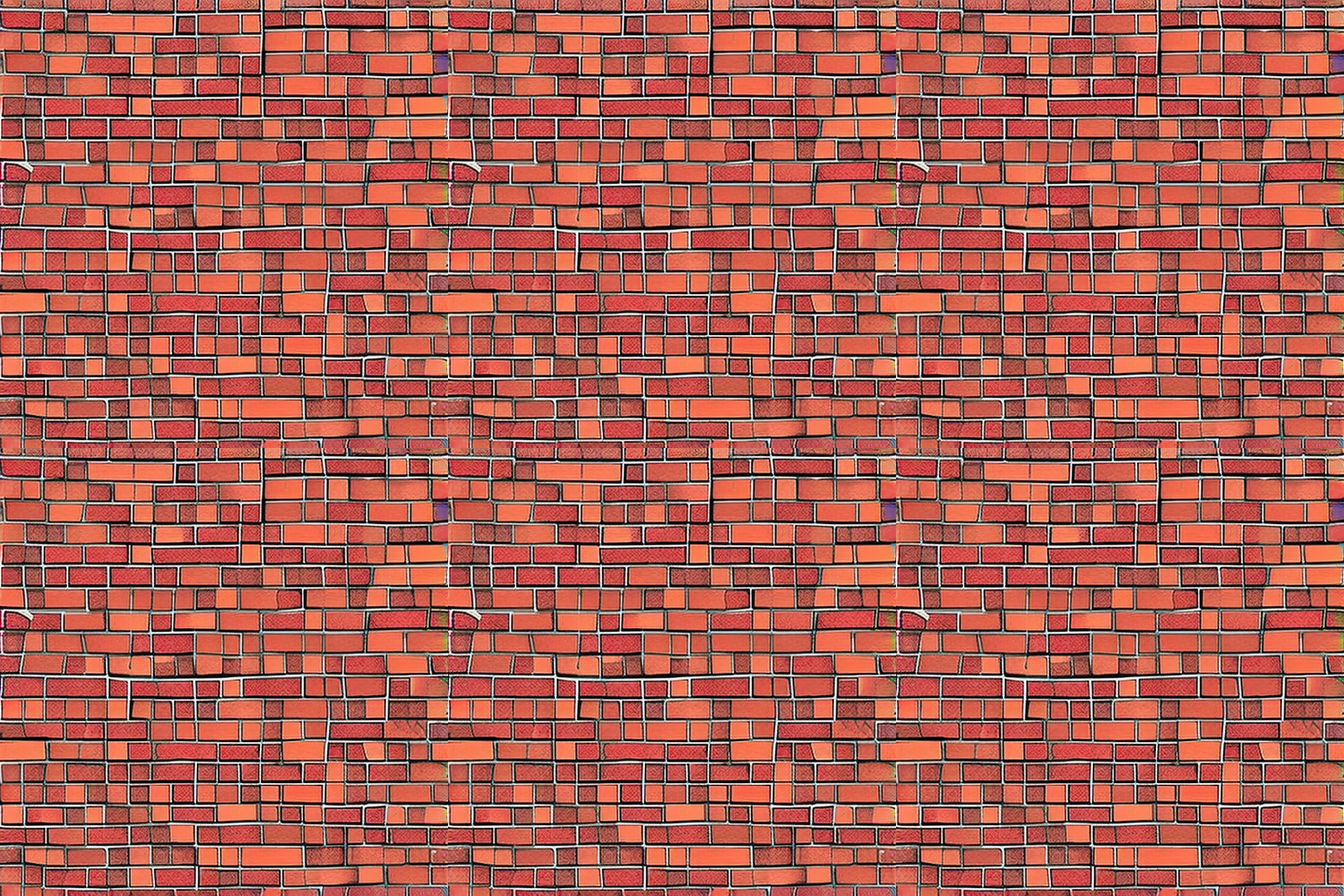} \\

        \end{tabular}
\caption{\textbf{Qualitative comparison of self-tiling.} The leftmost column shows the tiling constraints, followed by results from Tiled Diffusion, STI, and AT. The first two rows demonstrate 1x3 horizontal tiling of landscapes, while the last row shows 2x3 tiling of a texture. In the landscape examples (rows 1-2), our method and AT maintain logical continuity, while STI shows misalignments and artifacts at connection region. For the texture example (row 3), all methods perform comparably well due to the lack of complex high-level structure.}
\label{fig:qualitative:self}
\end{figure*}

\newlength{\selfwidthcc}
\setlength{\selfwidthcc}{3.0cm}

\newlength{\selfwidthrc}
\setlength{\selfwidthrc}{6.75cm}

\begin{figure*}[ht]
    
    \centering
        \centering
        \setlength{\tabcolsep}{4pt} %
        \begin{tabular}{ccc}
             & Tiled Diffusion & STI \\[0.1cm]
             
            \includegraphics[width=\selfwidthcc]{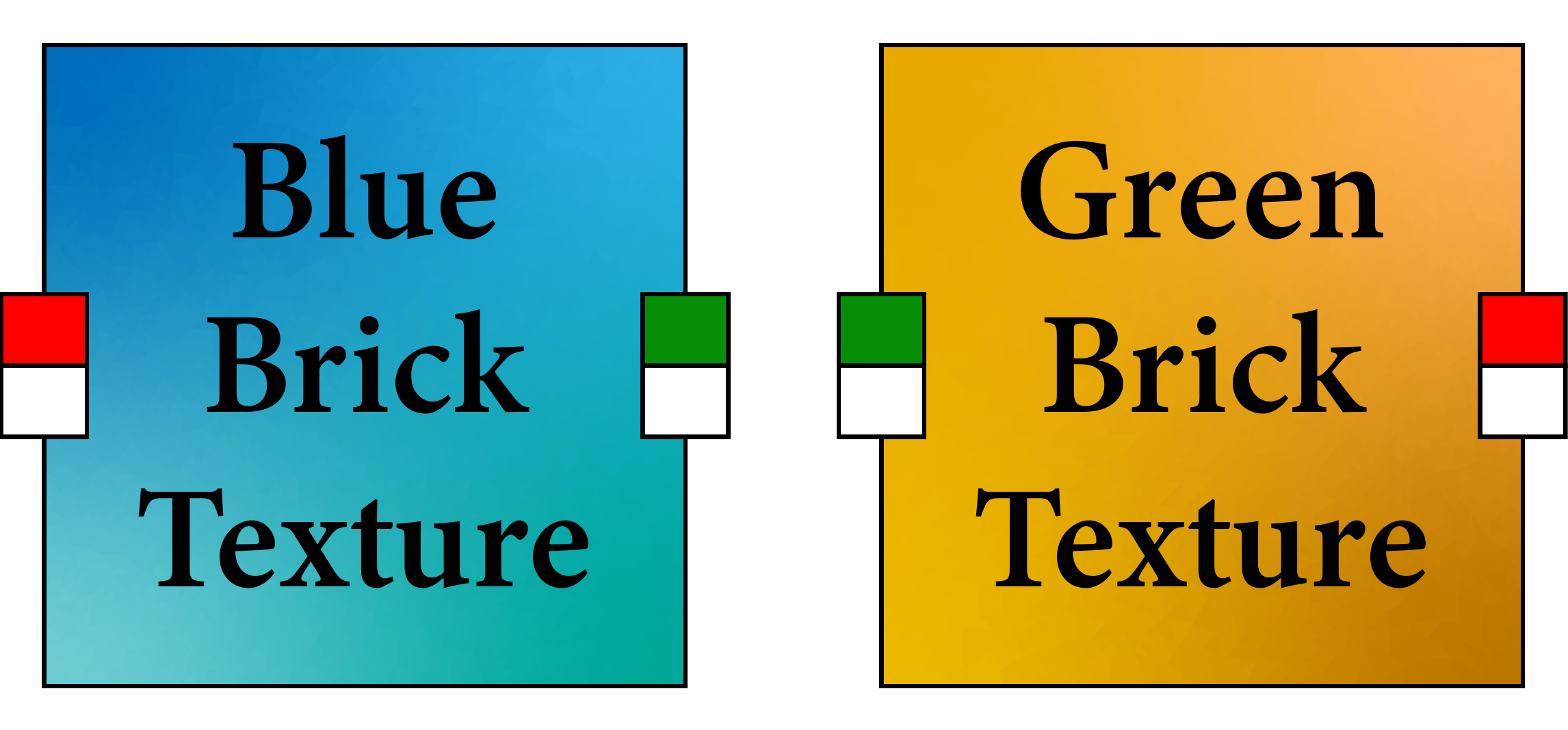} &
            \includegraphics[width=\selfwidthrc,frame]{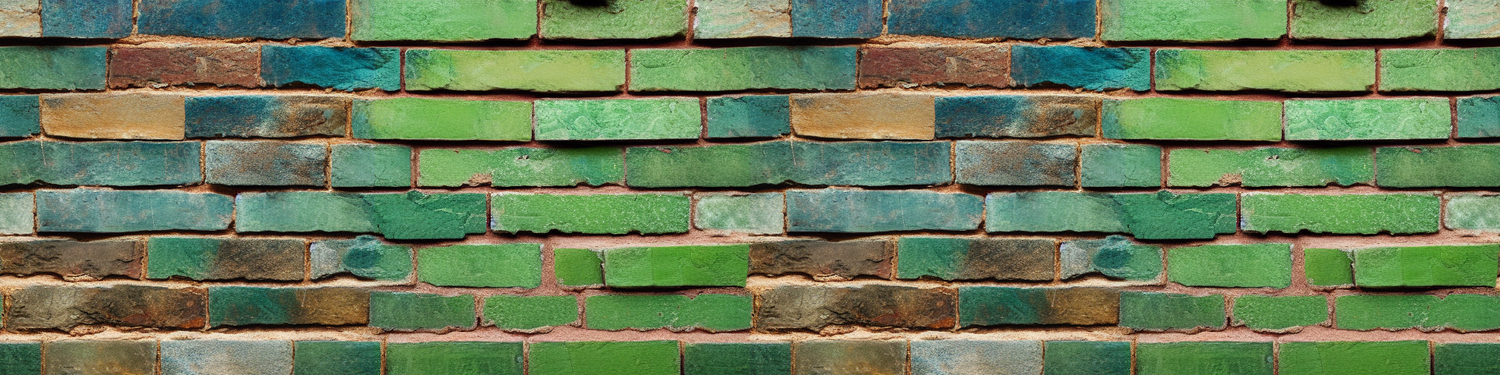} &
            \includegraphics[width=\selfwidthrc,frame]{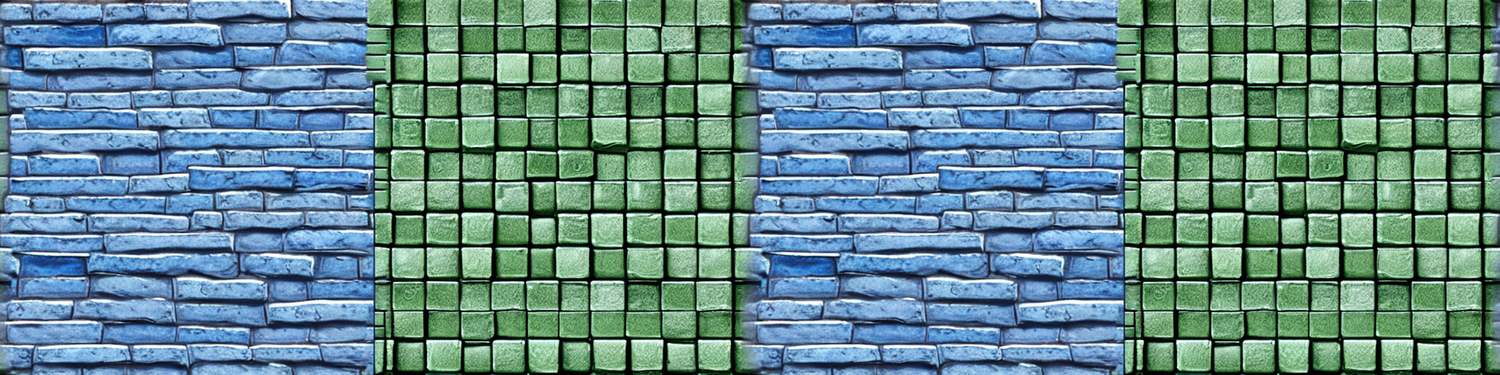} \\

            \includegraphics[width=\selfwidthcc]{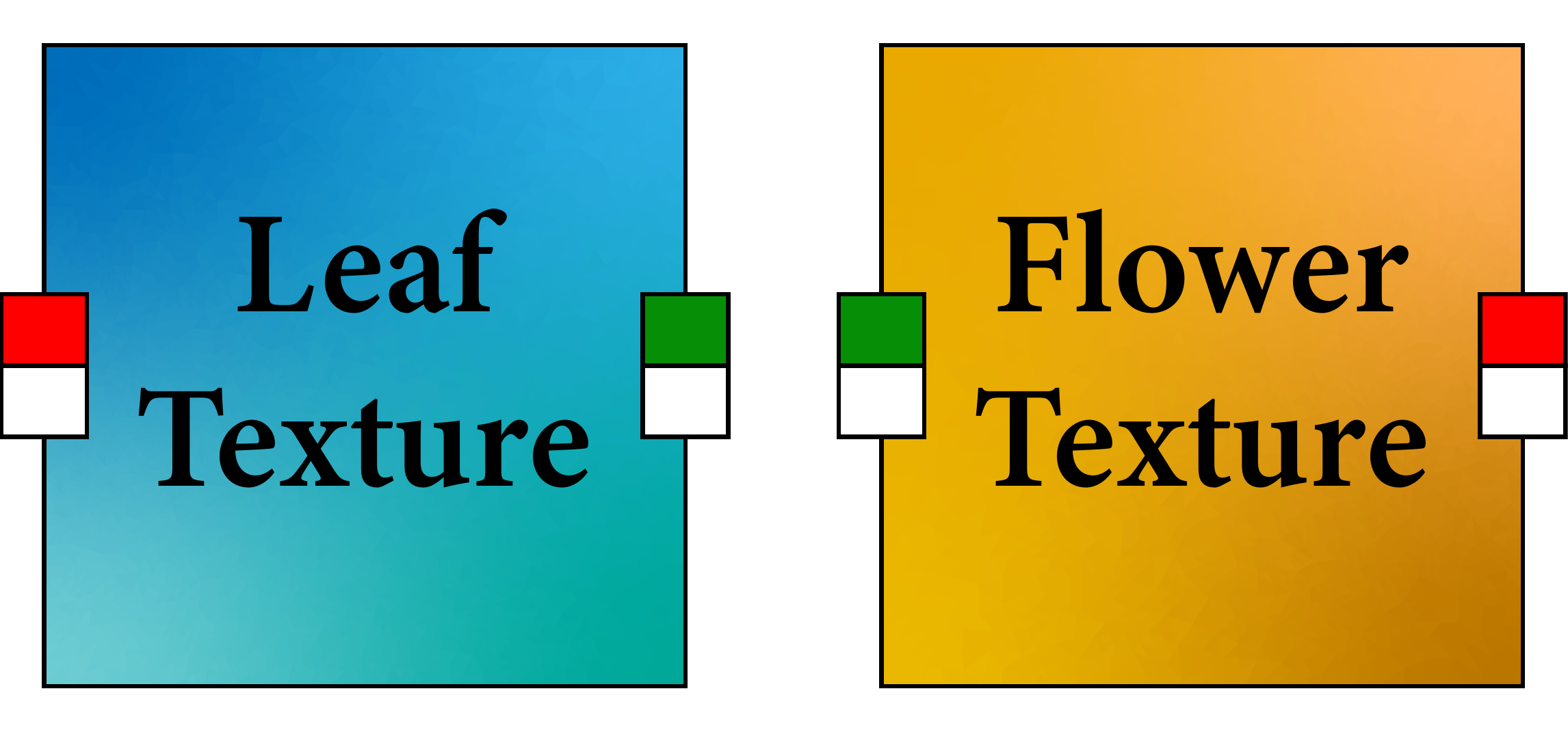} &
            \includegraphics[width=\selfwidthrc,frame]{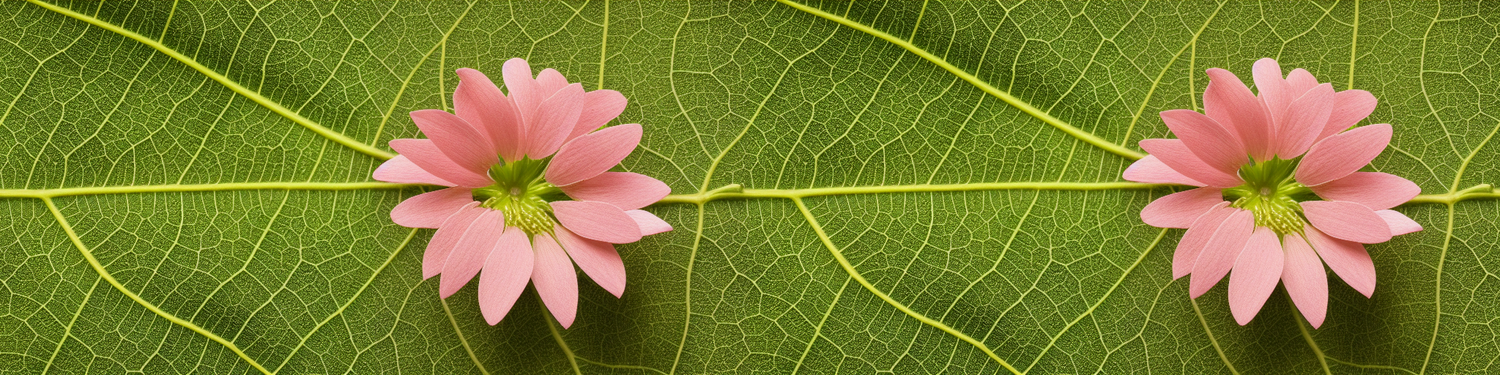} &
            \includegraphics[width=\selfwidthrc,frame]{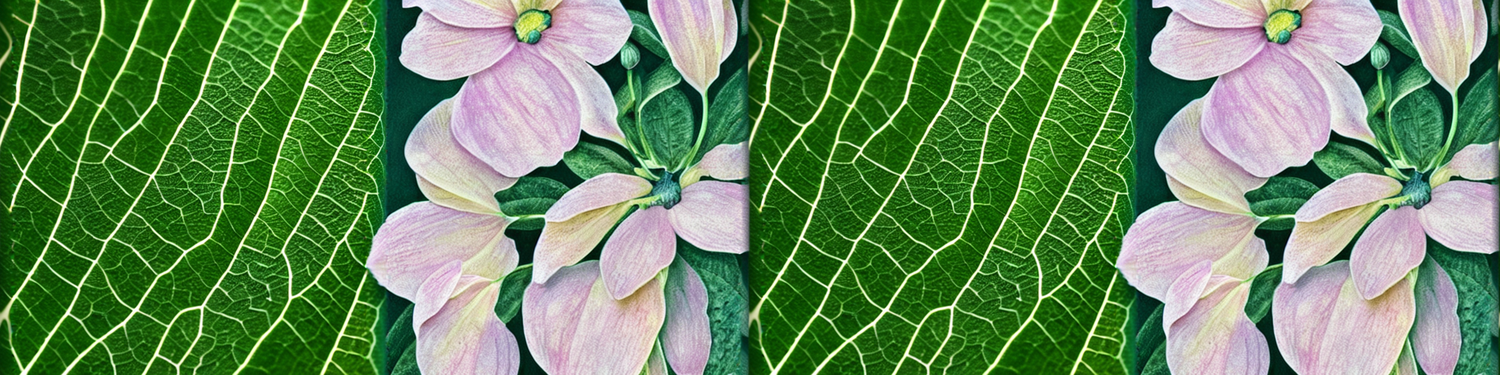} \\

            \includegraphics[width=\selfwidthcc]{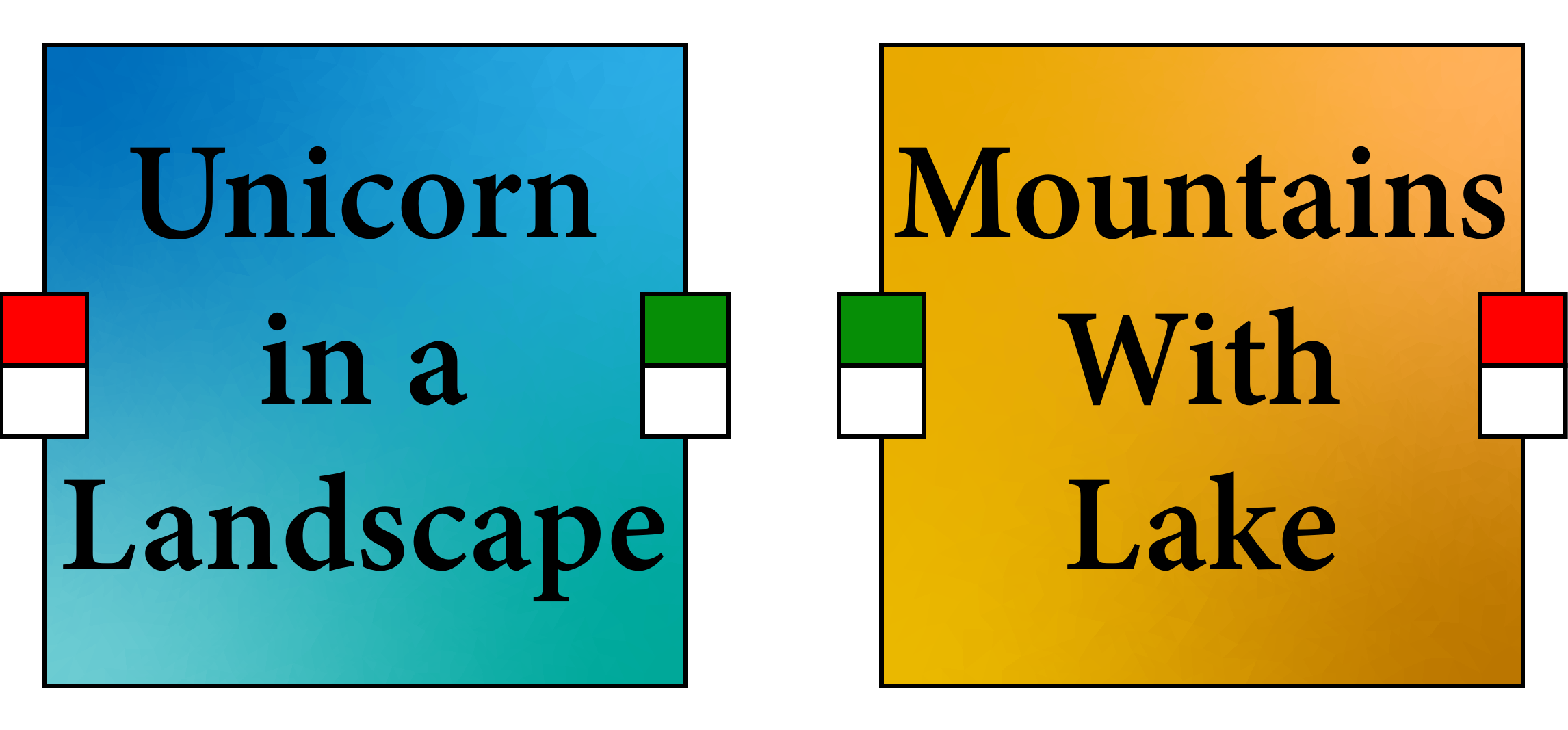} &
            \includegraphics[width=\selfwidthrc,frame]{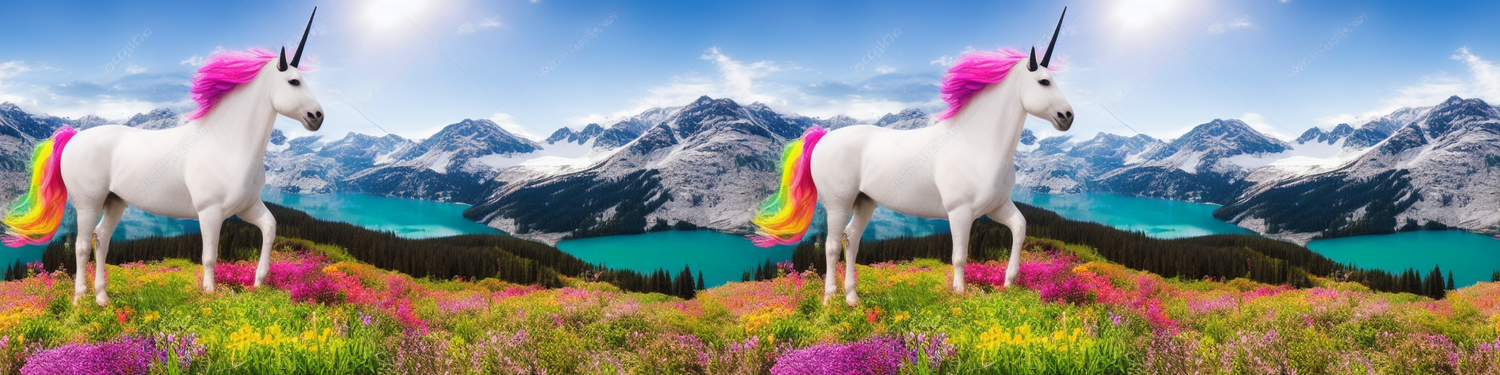} &
            \includegraphics[width=\selfwidthrc,frame]{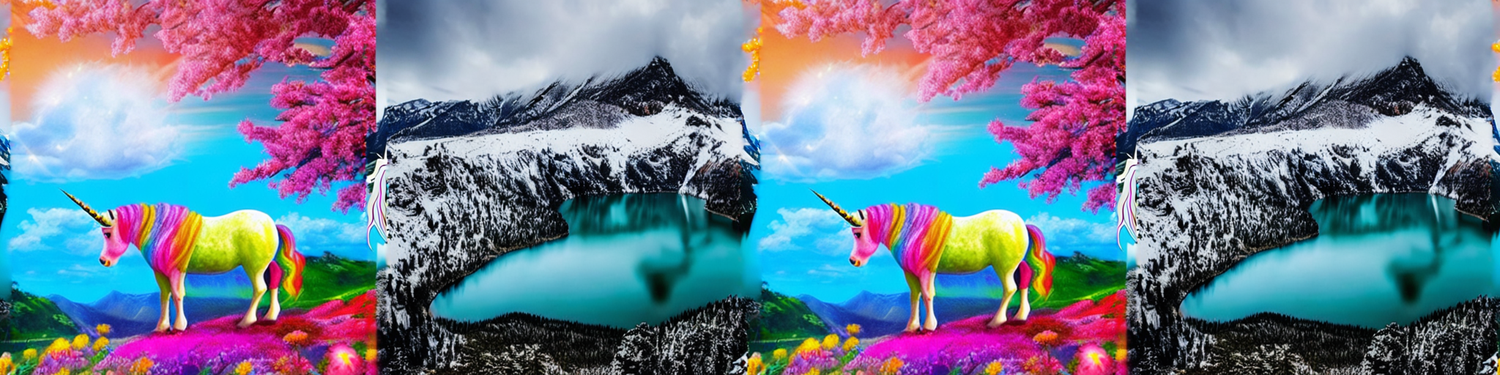} \\
            
        \end{tabular}
    \caption{
    \textbf{Qualitative comparison of one-to-one tiling.} The leftmost column displays the tiling constraints, followed by results from Tiled Diffusion and STI, respectively. All outputs are tiled 1x4. Our Tiled Diffusion method demonstrates superior coherence and seamless transitions between tiles. In contrast, the STI method, despite using a relatively large 80-pixel wide connection region between 512x512 pixel images, shows noticeable discontinuities and struggles to maintain consistent content across tile borders.
    }
    \label{fig:qualitative:one-to-one}
\end{figure*}

%% file: figures/qualitative/many-to-many.tex
\begin{figure*}[t]
    \centering
    \includegraphics[width=\textwidth]{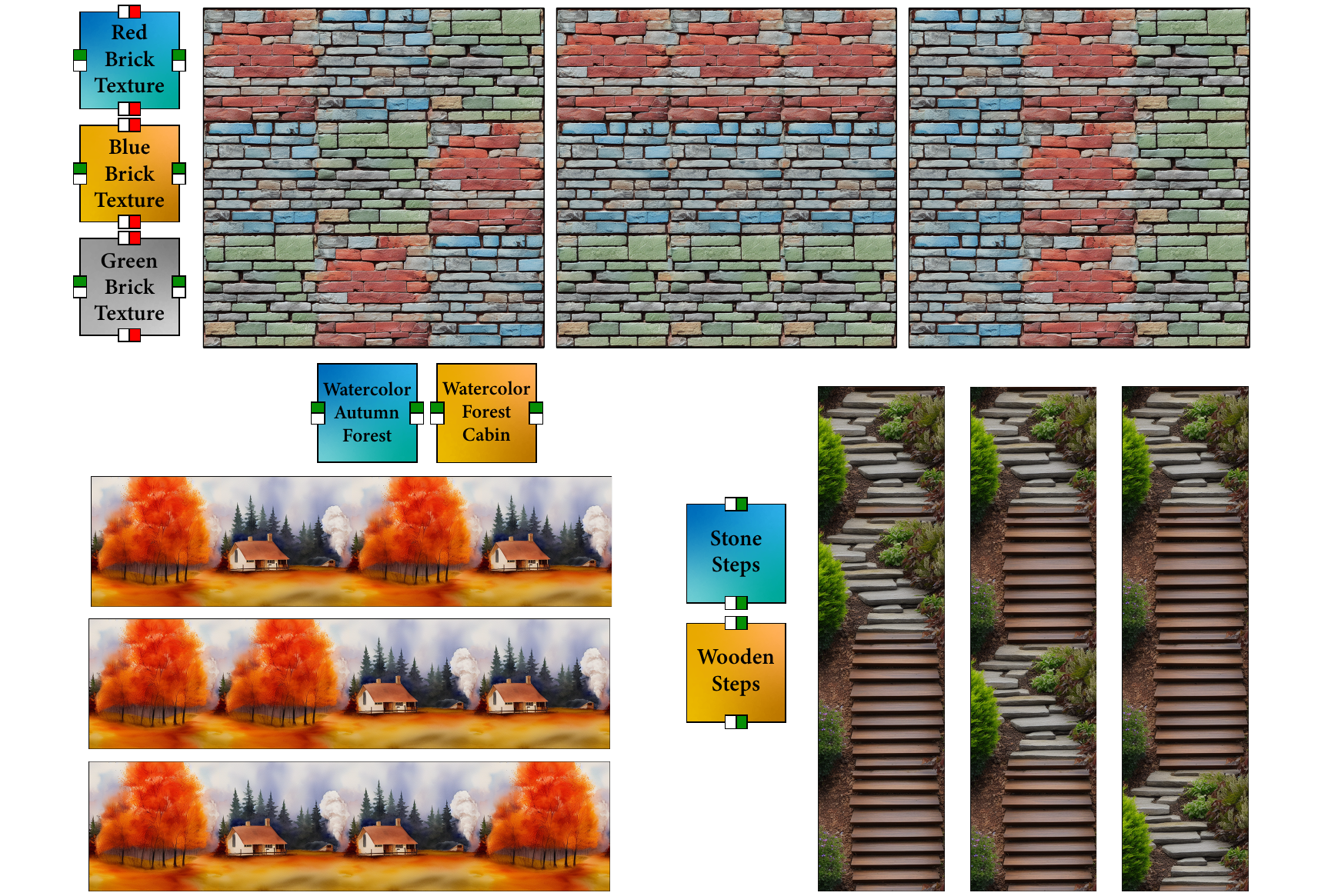}
    \caption{\textbf{Qualitative demonstration of Many-to-Many tiling.} Many-to-many connections enable several connection patterns between the resulting images. Top: 3x3 tiling of bricks. Bottom left: 1x4 horizontal tiling of watercolor forest paintings. Bottom right: 4x1 vertical tiling of different steps.}
\label{fig:qualitative:many-to-many}
\end{figure*}

%% file: sec/6_limitations.tex
\section{Limitations}
\label{sec:limitations}
Our method has a limitation with cross-axis connections (X-axis to Y-axis or vice versa), where small artifacts may appear in connecting regions. This stems from latent space rotations not perfectly aligning with pixel space rotations (after decoding), as the context copying during tiling constraint application requires rotation for cross-axis alignment which may not perfectly translate to pixel space. While these artifacts are typically minor and often imperceptible, future work could explore more sophisticated rotation techniques or post-processing steps to refine cross-axis connections. A detailed overview is provided in the supplementary material.

%% file: sec/7_conclusion.tex
\section{Conclusion}
\label{sec:conclusion}
We introduced Tiled Diffusion, a novel approach that extends diffusion models to accommodate complex tiling scenarios - self-tiling, one-to-one, and many-to-many. Our key innovations include a flexible tiling constraint and a similarity constraint, that ensure global structure consistency and eliminate artifacts in complex tiling scenarios. We also introduced a tiling score to measure tiling quality.
Our evaluations demonstrate the effectiveness of our approach, with superior performance across various metrics and diverse scenarios. We explored compelling applications in seamlessly tiling existing images, tileable texture generation, and 360\textdegree{} synthesis, showcasing the potential impact on texture creation, digital art, and other fields.

%% file: sec/8_ack.tex
\section{Acknowledgments}
\label{sec:ack}
This work was supported in part by the Israel Science Foundation (grant No. 1574/21).